\documentclass[10pt,twocolumn]{article}

\usepackage{graphicx}
\graphicspath{ {fig/} }
\usepackage{times}
\usepackage{graphicx} 
\usepackage{subfigure} 

\usepackage{natbib}
\usepackage{enumitem}

\usepackage{algorithm}
\usepackage{algorithmic}

\usepackage{amsmath}
\usepackage{amssymb}

\usepackage[toc,page]{appendix}
\numberwithin{equation}{section}

\usepackage{tikz}
\usetikzlibrary{er,positioning,shapes,snakes,shadows,arrows, topaths,matrix,arrows} 

\makeatletter
\pgfkeys{/pgf/.cd,
  parallelepiped offset x/.initial=2mm,
  parallelepiped offset y/.initial=2mm
}
\pgfdeclareshape{parallelepiped}
{
  \inheritsavedanchors[from=rectangle] 
  \inheritanchorborder[from=rectangle]
  \inheritanchor[from=rectangle]{north}
  \inheritanchor[from=rectangle]{north west}
  \inheritanchor[from=rectangle]{north east}
  \inheritanchor[from=rectangle]{center}
  \inheritanchor[from=rectangle]{west}
  \inheritanchor[from=rectangle]{east}
  \inheritanchor[from=rectangle]{mid}
  \inheritanchor[from=rectangle]{mid west}
  \inheritanchor[from=rectangle]{mid east}
  \inheritanchor[from=rectangle]{base}
  \inheritanchor[from=rectangle]{base west}
  \inheritanchor[from=rectangle]{base east}
  \inheritanchor[from=rectangle]{south}
  \inheritanchor[from=rectangle]{south west}
  \inheritanchor[from=rectangle]{south east}
  \backgroundpath{
    \southwest \pgf@xa=\pgf@x \pgf@ya=\pgf@y
    \northeast \pgf@xb=\pgf@x \pgf@yb=\pgf@y
    \pgfmathsetlength\pgfutil@tempdima{\pgfkeysvalueof{/pgf/parallelepiped offset x}}
    \pgfmathsetlength\pgfutil@tempdimb{\pgfkeysvalueof{/pgf/parallelepiped offset y}}
    \def\ppd@offset{\pgfpoint{\pgfutil@tempdima}{\pgfutil@tempdimb}}
    \pgfpathmoveto{\pgfqpoint{\pgf@xa}{\pgf@ya}}
    \pgfpathlineto{\pgfqpoint{\pgf@xb}{\pgf@ya}}
    \pgfpathlineto{\pgfqpoint{\pgf@xb}{\pgf@yb}}
    \pgfpathlineto{\pgfqpoint{\pgf@xa}{\pgf@yb}}
    \pgfpathclose
    \pgfpathmoveto{\pgfqpoint{\pgf@xb}{\pgf@ya}}
    \pgfpathlineto{\pgfpointadd{\pgfpoint{\pgf@xb}{\pgf@ya}}{\ppd@offset}}
    \pgfpathlineto{\pgfpointadd{\pgfpoint{\pgf@xb}{\pgf@yb}}{\ppd@offset}}
    \pgfpathlineto{\pgfpointadd{\pgfpoint{\pgf@xa}{\pgf@yb}}{\ppd@offset}}
    \pgfpathlineto{\pgfqpoint{\pgf@xa}{\pgf@yb}}
    \pgfpathmoveto{\pgfqpoint{\pgf@xb}{\pgf@yb}}
    \pgfpathlineto{\pgfpointadd{\pgfpoint{\pgf@xb}{\pgf@yb}}{\ppd@offset}}
  }
}
\makeatother

\begin{document} 

\title{Towards universal neural nets: Gibbs machines and ACE.}
\author{Galin Georgiev \\ GammaDynamics, LLC\footnote{galin.georgiev@gammadynamics.com}}
\date{}
\maketitle

\vskip 0.3in

\begin{abstract} 
We study from a physics viewpoint a class of generative neural nets, \emph{Gibbs machines}, designed for gradual learning. While including variational auto-encoders, they offer a broader universal platform for incrementally adding newly learned features, including  physical symmetries. Their direct connection to statistical physics and information geometry is established. A variational Pythagorean theorem justifies invoking the exponential/Gibbs class of probabilities for creating brand new objects.  Combining these nets with classifiers, gives rise to a brand of universal generative neural nets - stochastic auto-classifier-encoders (ACE). ACE have state-of-the-art performance in their class, both for classification and density estimation for the MNIST data set. 
\end{abstract}

\tableofcontents
\section{Introduction.}
\label{Introduction}

\subsection{Universality.}
\label{Universality}
We buck the recent trend of building highly specialized neural nets by exploring nets which accomplish multiple tasks without compromising performance. An \textit{universal} net can be tentatively described as one which, among other things: i) works for a variety of applications, i.e. visual recognition/reconstruction, speech recognition/reconstruction, natural language processing, etc; ii) performs various tasks: classification, generation, probability density estimation, etc; iii) is self-contained, i.e., does not use specialized external machine learning methods; iv) is biologically plausible.

\subsection{Probabilistic and quantum viewpoint on generative nets.}
\label{Probabilistic and}
The input of a  neural net is typically  a $P \times N$ \emph{data matrix} $\mathbf X$. Its row-vectors  $\{\mathbf{x}_{\mu}\}_{\mu=1}^P$ span the space of {\emph observations}, its column-vectors $\{ \mathbf{x}_i\}_{i=1}^N$, where $1...N$ can for example enumerate the pixels on a screen, span the space of \emph{observables}. The net is then asked to perform classification, estimation, generation, etc, tasks on it.   In \emph{generative} nets, this is  accomplished by randomly generating $L$ latent observations  $\{\mathbf{z}_{\mu}^{(\kappa)} \}_{\kappa=1}^L$ for every observation $\mathbf{x}_{\mu}$. This induced ``uncertainty'' of the $\mu$-th state is modeled by a \emph{model} conditional density $p(\mathbf{z|x}_{\mu})$. It is the copy-cat, in imaginary time/space, of the (squared) wave function from quantum mechanics\footnote{ Strictly speaking, we will employ unbounded densities and hence  stochastic analysis formalism and its centerpiece - the diffusion equation. But they are formally equivalent to the quantum-mechanical formalism and its centerpiece - the Schrodinger equation - in imaginary time/space coordinates.}, and fully describes the $\mu$-th conditional state $\left(\mathbf{x}_{\mu}, \{\mathbf{z}_{\mu}^{(\kappa)} \}_{\kappa=1}^L \right)$. In statistical  mechanics parlance, the \emph{latents} are fluctuating  microscopic variables, while the macroscopic observables are obtained from them via some aggregation. In the absence of  physical time, observations are thus interpreted as \emph{partial equilibria} of  independent small  parts of the expanded (by a factor of $L$) original data set. Simply put, every visible observations is surrounded by a ``cloud'' of virtual observations. Creating a new original observation amounts to nothing more than sampling from that cloud.

The quality of the model conditional density, or more generally - the model joint density  $p(\mathbf{z,x}_{\mu})$ - is judged by the ``distance'' from the implied marginal density $q(\mathbf{x})$ $:=\int p(\mathbf{x,z})d\mathbf{z}$ to the \emph{empirical} marginal density $r(\mathbf{x})$. This distance is  called \emph{cross-entropy} or  \emph{negative log-likelihood}:
\begin{align}
-\log \mathcal{L}(r||q) := \mathbb{E}_{r(\mathbf{x})}[- \log q(\mathbf{x})],
\label{1.-1}
\end{align}
where $\mathbb{E}_{r()}[.]$ is an expectation with respect to $r()$. Its minimization is the ultimate goal.

\subsection{Equilibrium setting. Gibbs machines.}
\label{Equilibrium setting}
The equilibrium i.e. small fluctuations viewpoint of statistical mechanics appears to have been originated by Einstein in a then-unpublished 1910 lecture \cite{Einstein10}. He used an  exponential  model density $p^{Exp}()$ in space and  derived the Brownian diffusion i.e., Gaussian  model density $p^G()$ in space/time.  They are special cases of a broad class of densities - \emph{Gibbs} or \emph{exponential} densities - which form the foundation of classic statistical mechanics. Gibbs densities are variational \emph{maximum-entropy} densities and hence optimal for modeling equilibria. They also offer a platform  for  adding incrementally new macroscopic descriptive variables, \cite{Landau80}, section 35. 

We argue in sub-sections \ref{Generative error}, \ref{Gibbs machines} that  Gibbs densities are also optimal for modeling \emph{fully-generative} equilibrium nets, and call those nets  \emph{Gibbs machines}. They were inspired by the first fully-generative nets - the \emph{variational} auto-encoders (VAE) \cite{Kingma14-1}, \cite{Rezende14} - and employ the same upper bound for the cross-entropy target (\ref{1.-1}). Like their physics counterparts, Gibbs machines offer a platform for mimicking the gradual nature of learning: already learned \emph{symmetry statistics} like   space/time symmetries, can be added incrementally and accelerate learning, sections \ref{Symmetries in}, \ref{Gibbs and}, \ref{Latent symmetry}.

\subsection{Observation entropies}
\label{Observation entropies}

Unlike equilibrium statistical mechanics, human data is decidedly non-equilibrium in nature and exhibits large fluctuations and non-Gaussian behavior. Quantifying non-Gaussianity and ``distance'' from equilibrium is not easy when dealing with  large number of observables $N$ and observations $P$. Luckily, there is a one-dimensional proxy for  non-Gaussianity of a multi-dimensional data set: the non-Gaussianity of the negative Gaussian log-likelihoods  $ \{- \log p^G(\mathbf{z_{\mu}}) \}_{\mu}$. Here, $ - \log p^G(\mathbf{z_{\mu}})$  $=\frac{1}{2}\mathbf{z}_{\mu}\mathbf{C}(\mathbf{z})^{ -1}\mathbf{z}_{\mu}$ $+const$, with a multivariate Gaussian  $\mathcal{N}_{N_{lat}}(0,\mathbf{C}(\mathbf{z}) )$ as model density and $\mathbf{C}(\mathbf{z})$ the empirical covariance\footnote{If the latent observations $\{\mathbf{z}_{\mu}\}_{\mu=1}^P$ come from an $N_{lat}$-dimensional Gaussian distribution, the density of these negative Gaussian log-likelihoods  is proportional to the familiar $F(N_{lat}, P-N_{lat})$ density \cite{Mardia79}, sections 1.8, 3.5. 
For the typical case $P-N_{lat} \rightarrow \infty$, it is proportional to the chi-squared density $\chi_{N_{lat}}^2()$, which in turn converges to a rescaled Gaussian $\mathcal{N}(0,1)$, as $N_{lat} \rightarrow \infty$.}. 

In a bold and counter-intuitive re-read of Boltzmann's statistical mechanics, Einstein interpreted the log-likelihoods $ \{\log p^G(\mathbf{z_{\mu}}) \}_{\mu}$ as \emph{observation entropies}  \cite{Einstein10}. We will denote the Einstein entropy\footnote{In order for the negative logs to be thought of as entropies, a large positive constant is added. It should be clear from the context, but we nevertheless use for clarity a superscript, to distinguish the Einstein entropy $\mathcal{S}^E(.)$ of an observations, from the standard Boltzmann entropy $\mathcal{S}(.)$ of a probability density, introduced in sub-section \ref{Conditional densities}.} of an observation $\mathbf{z}_{\mu}$  by:
\begin{align}
\mathcal{S}^E(\mathbf{z}_{\mu}) := \log p^G(\mathbf{z_{\mu}}) + const \geq 0. 
\label{1.0}
\end{align}
The sub-section \ref{Probabilistic and}  viewpoint of an observation, as a partial equilibrium of multiple virtual  observations, fits right into Einstein's paradigm: entropy is defined in a classical Boltzmann fashion, but on a cloud of virtual observations. The visible observation stands out as the one with a locally maximum entropy. 

Observation entropies are central to modern theory of fluctuations  \cite{Landau80}, chapter 12. 
They also have an elegant linear-algebraic incarnation in the singular value decomposition of the data matrix $\mathbf X$ (sub-section \ref{Non-generative ACE}). In addition, their second moment $ \mathbb{E}_r\left[ (\log p^G(\mathbf{z_{\mu}}) )^2 \right]$ is proportional to the multi-variate \emph{kurtosis}, measuring the ``fatness'' of the probability density of $\{\mathbf{z}_{\mu}\}$.
 
\subsection{The equilibrium curse. Intricates.}
\label{The curse}

\begin{figure}[!ht]
\vskip 0.2in
\begin{minipage}{0.5\columnwidth}
\includegraphics[width=\columnwidth]{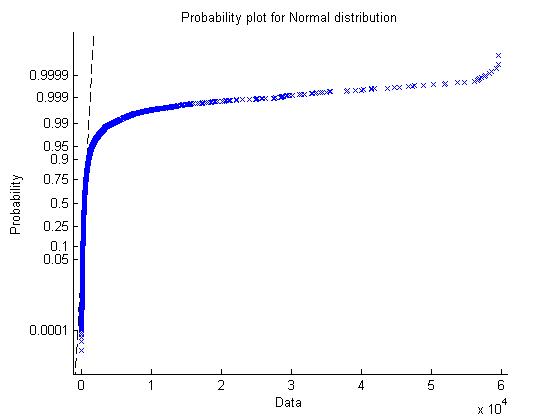}
\end{minipage}%
\begin{minipage}{\columnwidth}
\includegraphics[width=0.5\columnwidth]{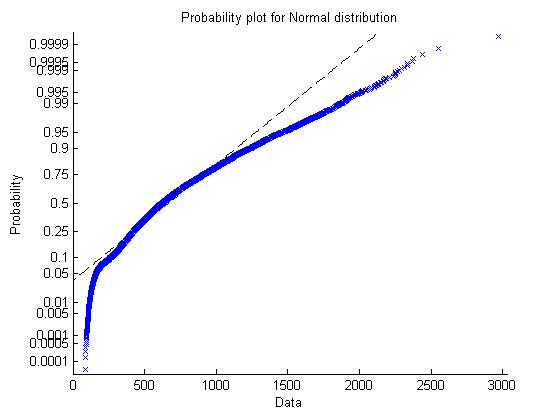}
\end{minipage}  \hfill
\caption{Q-Q plots against a Gaussian of the density of the negative log-likelihoods $\{ - \log p^G(\mathbf{z_{\mu}})\}_{\mu}$ of the  10000  MNIST test observations in layer  2 of a 5-layer standard feed-forward classifier net, see right branch of Figure \ref{Fig1.4} and Appendix \ref{Software and} for implementation details. Layer sizes are 784-700-700-700-10. Learning rate = 0.0015, decay = 500 epochs, batch size = 10000. For the right plot, dropout is 0.2 in input layer and 0.5 in hidden layers. As an exception from  the rules in Appendix \ref{Software and}, a tanh() activation function is used in the first hidden layer. 
\textbf{Left.} No dropout and no non-linearity: highly non-Gaussian.  
\textbf{Right.} With dropout and non-linearity: severely Gaussianized, especially for the intricates towards the right (see text).}
\label{Fig1.1}
\end{figure}

Unfortunately, some of the key features of modern  neural nets, like non-linear activation functions and dropout \cite{Srivastava14}, come  at the high price of  \emph{Gaussianizing} the data set, i.e. lead to higher-entropy, less informative configurations. The right \emph{quantile-quantile} (Q-Q) plot in Figure \ref{Fig1.1} shows the Gaussianization effect of non-linearities and dropout for the MNIST data set \cite{LeCun98}. Bounded non-linearities Gaussianize because they are compressive in nature and ``straighten out'' the unlikely (with small entropies) observations, which we  refer to as \emph{intricates}. Dropout Gaussianizes because it drops latent  variables and thus decreases the  kurtosis.

\begin{figure}[!ht]
\begin{minipage}{0.5\columnwidth}
\includegraphics[width=\columnwidth]{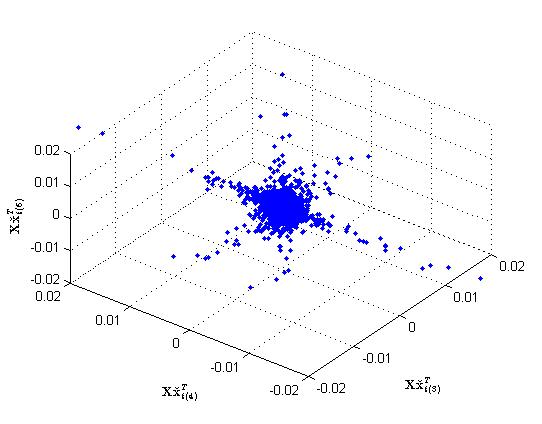}
\end{minipage}%
\begin{minipage}{0.5\columnwidth}
\includegraphics[width=\columnwidth]{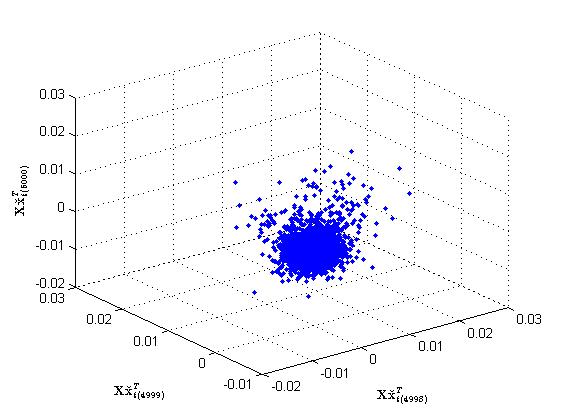}
\end{minipage}  \hfill
\caption{\textbf{Left.} The first 5000 MNIST training images, projected on three of the least likely, i.e. most intricate  conjugate images, ranked \#3, \#4, \#6 in ascending order of Einstein entropy $ \mathcal{S}^E(\mathbf{x}_{\mu})$, (\ref{1.0}). This is a highly non-Gaussian 3-dimensional distribution.  
\textbf{Right.} The same MNIST images, projected on the three most likely conjugate images, ranked \# 4998, \#4999, \#5000 in Einstein entropy. Much more Gaussian-looking.}
\label{Fig1.2}
\end{figure}

\begin{figure}[!ht]
\begin{minipage}{\columnwidth}
\includegraphics[width=\columnwidth]{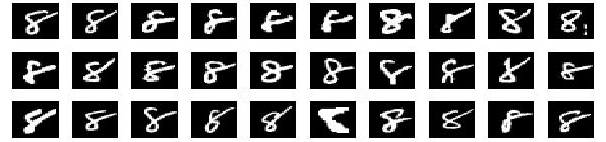}
\end{minipage}
\vskip 0.2in
\begin{minipage}{\columnwidth}
\includegraphics[width=\columnwidth]{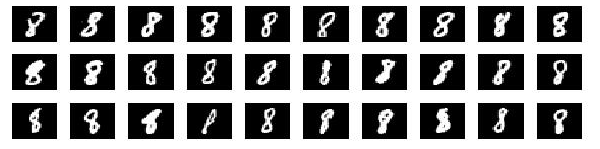}
\end{minipage}  \hfill
\caption{ \textbf{Top.} The 30 lowest-entropy MNIST training images, using the Einstein entropy  $\mathcal{S}^E(\mathbf{z}_{\mu})$, (\ref{1.0}), from the class corresponding to the  digit 8. They are quite intricate indeed.  \textbf{Bottom.} The 30 highest-entropy MNIST training images from the same class. Much more vanilla-looking.}
\label{Fig1.3}
\end{figure}

It is precisely the intricates, which - because of their low entropy and extreme non-Gaussianity, see Figure \ref{Fig1.2} - are  ideal candidates for ``feature vectors''   in classification tasks \cite{Hyvarinen09}, section 7.9, 7.10. Their conjugates are then the ``receptive fields'' or ``feature detectors''\footnote{Recall that, for a given row-vector observation $\mathbf{x}_{\mu}$, its \emph{conjugate} is $\check{\mathbf{x}}_{\mu}$ $= \mathbf{x}_{\mu} \mathbf{C}^{-1}$, where $\mathbf{C}$ is the covariance matrix. Up to a constant, the Gaussian negative log-likelihood is thus the inner product of an observation and its conjugate, in the standard Euclidean metric: $ - \log p^G(\mathbf{x_{\mu}})$  $=\frac{1}{2}<\check{\mathbf{x}}_{\mu},\mathbf{x}_{\mu}>$.} - see open problem \ref{Intricates} in section \ref{Open problems}. We show on the top (respectively, bottom) plot in Figure \ref{Fig1.3} the 30 least (respectively, most) likely images in MNIST, in ascending order of their entropy  $\mathcal{S}^E$, from the class corresponding to the  digit 8.

\subsection{Symmetries in the latent manifold.}
\label{Symmetries in}

\begin{figure}[!h]
\vskip 0.2in
\includegraphics[width=\columnwidth]{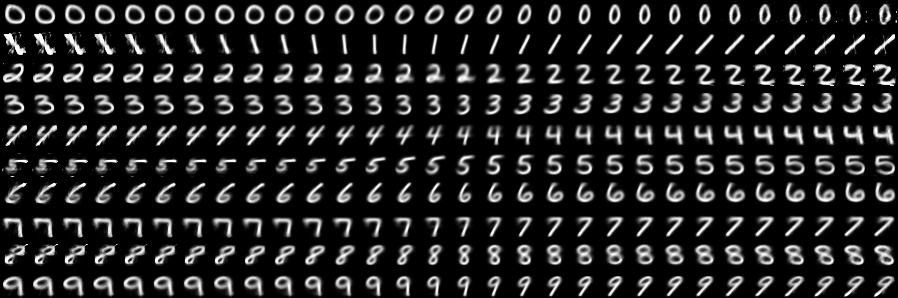}
\caption{Dominant dimension for each of MNIST ten classes, with each row corresponding to a separate class. While rotational symmetry dominates most classes, size i.e. scaling symmetry, clearly dominates the class of digit 5. 
The net is an ACE in  creative regime as in sub-section \ref{Fully-generative nets}, with an equally spaced  deterministic grid in the latent layer $\{\sigma_s \}_{s= 1}^{30}$, $-6 \leq \sigma_s \leq 6$. Layer sizes 784-700-(1 x10)-(700x10)-(784x10) for the AE branch and 784-700-700-700-10 for the C branch, Figure \ref{Fig1.4} and Appendix \ref{Software and}, learning rate = 0.0002, decay = 500 epochs, batch size = 1000.   }
\label{Fig1.5}
\end{figure}

When the dimensionality $N_{lat}$ of the ACE latent layer is low, traversing the latent dimension in some uniform fashion describes the latent manifold for a given class. Figure \ref{Fig1.5} shows the dominant dimension for each of the 10 classes in MNIST. This so-called \emph{manifold learning} by modern feed-forward nets was pioneered by the \emph{contractive} auto-encoders (CAE) \cite{Rifai12}. A symmetry in our context is, loosely speaking, a one-dimensional parametric transformation, which leaves the cross-entropy unchanged. In probabilistic terms, this is equivalent to the existence of a one-parametric density, from which ``symmetric'' observations are sampled, see (\ref{1.2}) below. Nets currently learn  symmetries from the training data, after it is artificially \emph{augmented},  e.g. by adding rotated, translated, rescaled, etc,  images,   in the case of visual recognition. But once a symmetry is learned, it does not make sense to re-learn it for every new data set. 

We hence propose the \emph{reverse} approach: add the symmetry explicitly to the latent  layer, alongside its Noether invariant, \cite{Gelfand63}. 
Take for example translational symmetries in a two-dimensional system with coordinates $(z^{(h)},z^{(v)})$ $\in \mathbb{R}^2$. They imply the conservation of the horizontal and vertical \emph{momenta} $(-i\hslash\partial/\partial z^{(h)}, -i\hslash\partial/\partial z^{(v)})$    $=(p^{(h)}, p^{(v)})$ $\in \mathbb{R}^2$ and a quantum mechanical wave function $\sim e^{ \frac{i}{\hslash} p^{(h)} (z^{(h)} - h) + \frac{i}{\hslash} p^{(v)} (z^{(v)} - v)}$, where   $(h,v)$ are offsets, \cite{Landau77}, section 15. After switching to imaginary time/space as in sub-section \ref{Probabilistic and}, and setting $\hslash=1$, the corresponding conditional model density for a given observation/state $\mu$ is:
\begin{align}
p(\mathbf{z|x}_{\mu})  \sim  e^{-2p^{(h)}_{\mu} |z^{(h)} - h_{\mu}| -2p^{(v)}_{\mu} |z^{(v)} - v_{\mu}|},
\label{1.2}
\end{align}
i.e. a two-dimensional Laplacian which fits \footnote{ Technically, Laplacian is not in the exponential class, but it is a sum of two exponential densities in the domains $(-\infty, \mu)$, $[\mu, \infty)$ defined by its mean $\mu$, and those densities are in the exponential class in their respective domains. Laplacian  is biologically-plausible because it is a  bi-product of squaring Gaussians.} in the Gibbs machine paradigm (\ref{2.11}). We demonstrate in section \ref{Latent symmetry} how to build-in translational, scaling and rotational symmetry in a net, by computing the \emph{symmetry statistics} like $\{h_{\mu}, v_{\mu}\}$ explicitly  and estimating the  invariants with the rest of the net parameters. In general, they have to be refined via an optimization, as e.g. in \cite{Jadeberg15}. For more details, see \cite{Georgiev15-2}.

\subsection{ACE.}
\label{ACE}

\begin{figure*}[!ht]
\begin{center}
\begin{tikzpicture}

\node[rectangle, text width = 0.4\columnwidth, text  centered,  draw ] 
	(Input) at (5,0) {\small \underline{1 \textbf{Input layer} }:\\Size = \# observables  $N$ };
\node[rectangle,  text width = 0.6\columnwidth, text centered, below  left =.1 in and -0.2in of Input,  draw] 
	(AE_EncoderHidden)  {\small \underline{2. \textbf{AE encoder hidden layer(s)} }: \\Size = $N_{enc}$ };
\node[parallelepiped,  text width = 0.5\columnwidth, parallelepiped offset y=0.1in, text  centered, below   = 0.2in  of AE_EncoderHidden,draw] 
	(AE_Latent/Feature)  {\small \underline{3. \textbf{AE latent hidden layer(s)}}: \\Size = $N_{AE~lat} \times N_C$} ;
\node[parallelepiped,  text width = 0.6\columnwidth, parallelepiped offset y=0.1in, text  centered, below  =0.2in   of AE_Latent/Feature, draw] 
	(AE_DecoderHidden)  {\small \underline{4. \textbf{AE decoder hidden layer(s)}}:  \\Size = $N_{AE~dec} \times N_C$} ;
\node[parallelepiped,  text width = 0.6\columnwidth,  parallelepiped offset y=0.1in, text  centered, below=0.2inof AE_DecoderHidden, draw] 
	(AE_Output)  {\small \underline{5. \textbf{AE output layer}}:\\Size = $N \times N_C$};

\path [->](Input.west) edge node {} (AE_EncoderHidden); 
\path [->](AE_EncoderHidden) edge node {} (AE_Latent/Feature);
\path [->](AE_Latent/Feature) edge node {} (AE_DecoderHidden); 
\path [->](AE_DecoderHidden) edge node {} (AE_Output);

\node[rectangle,  text width = 0.4\columnwidth,  text  centered, below right =.5in and -0.3in of Input, draw] 
	(C_DecoderHidden)  {\small \underline{2. \textbf{C hidden layer(s)}}:\\ Size = $N_{C~hidden}$} ;
\node[rectangle,  text width = 0.4\columnwidth,  text  centered, below=.5inof C_DecoderHidden, draw] 
	(C_Output)  {\small \underline{3. \textbf{C output layer}}: \\ Size = \# classes $N_C$ };

\path [->](Input.east) edge  node {} (C_DecoderHidden);  
\path [->](C_DecoderHidden) edge node {} (C_Output); 
	
\path [->](C_Output) edge [dashed] node {} (AE_Latent/Feature); 
\path [->](C_Output) edge [dashed] node {} (AE_DecoderHidden);
\path [->](C_Output) edge [dashed] node {} (AE_Output); 
		
\draw [->]  ([xshift=0.3in]C_Output.east) edge [bend right=0, thick, double]  node [rotate=270, midway, yshift = 0.5in] { $\begin{array}{l} \textbf{Back-propagation } \\ of  \min \left(- log \mathcal{L}_{ACE} + reg~constraints\right),    \\ \text{where, } \\ \qquad  \boxed{- log \mathcal{L}_{ACE} = - log \mathcal{L}_{AE} - log \mathcal{L}_{C} }  \end{array} $} ([xshift=0.in -0.5in + 0.6\columnwidth]Input.east);
         
\end{tikzpicture}
\caption{ACE architecture: $\mathbf{AE}$ stands for ``auto-encoder'', $\mathbf{C}$ stands for ``classifier''.  Training is supervised i.e. labels are used in the auto-encoder and each class has a separate decoder, with  unimodal sampling in the latent layer.  The sampling  during testing is instead from a mixture of densities $p(\mathbf{z}|\mathbf{x}_{\mu})$ $=\sum_{c=1}^{N_c} \omega_{\mu,c} p(\mathbf{z}|\mathbf{x}_{\mu}, c)$, with class probabilities  $\{\omega_{\mu,c} \}_{c=1}^{N_c}$ provided by the classifier, hence the dashed lines.} 

\label{Fig1.4}
\end{center}
\vskip -0.2in
\end{figure*}
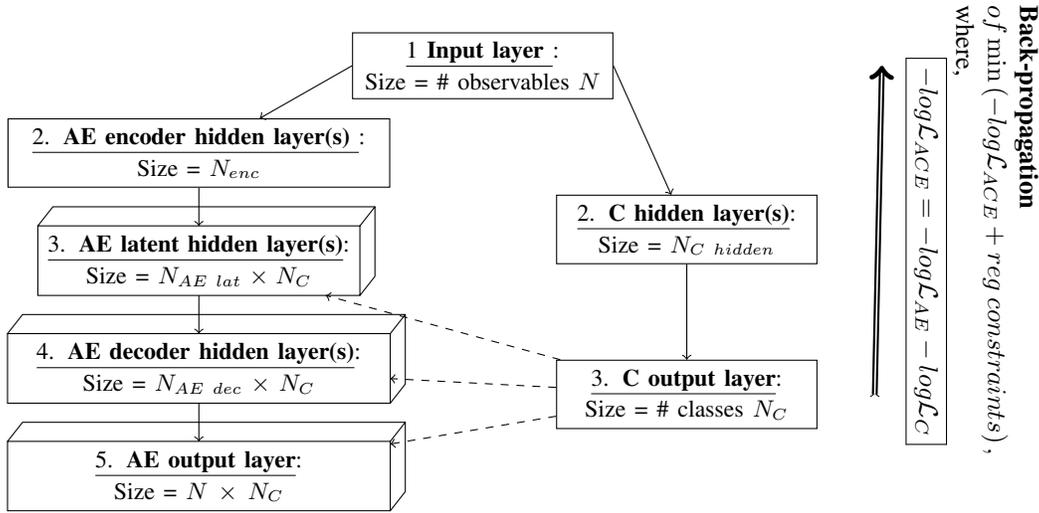

\subsubsection{Non-generative ACE.}
In order to preserve the non-Gaussianity of the data, and improve performance significantly along the way, we will combine classifiers with \emph{auto-encoders} - hence the name \emph{auto-classifier-encoder} (ACE).  Auto-encoders have a \emph{reconstruction} error in their cross-entropy optimization target and thus force the net to be more faithful to the raw data. ACE simultaneously classifies and  reconstructs, assuming an independence between the two, and hence additivity of  the respective cross-entropies:
\begin{align}
 - log \mathcal{L}_{ACE} = - log \mathcal{L}_{AE} - log \mathcal{L}_{C}.
\label{1.1}
\end{align}
In its first - non-generative - installment, ACE can do with a standard classifier and a shallow auto-encoder in the dual space of observations. It still beats handily the peers in its class, Figure \ref{Fig4.2}, right.\footnote{In its embryonic form, the shallow ACE seems to have first appeared in \cite{Le11}, with the purpose of  replacing  orthogonality constraints in   Independent Component Analysis.}

\subsubsection{Non-Gaussian densities.} 
In real-life data sets, the number of observations $P \rightarrow \infty$, while the dimension of observables $N$ is fixed. An universal net will hence tend to work better when the dimension of latent layers $N_{lat} \geq N$, i.e. have the so-called \emph{overcomplete} representation \cite{Coates11}.  When $N_{lat} >> N$, for any given $N$-dimensional observation $\mathbf{x}_{\mu}$, only a small number of latents $\{z_{\mu j} \}_{j=1}^{N_{lat}}$ deviate significantly from zero. For these \emph{sparse representations}, sampling  from high-entropy Gaussian-like densities, as on the right plot of Figure 2, is  flawed. Sampling instead from ``fat-tail'' densities offers a significant performance improvement for MNIST, Figure \ref{Fig4.3}, right. As in mathematical finance, stochastic volatility and jumps are arguably the first natural source of non-Gaussianity, and are almost fully-tractable. The \emph{q-Gibbs machines} offer another  venue, sub-section \ref{Gibbs and}. 

\subsubsection{Generative ACE.} 
An even greater issue for current nets is the spontaneous ``clumping'' or clusterization which is prevalent in real-life data sets. Statistical mechanics deals with it by introducing \emph{higher-hierarchy} densities which are conditional on low-hierarchy densities. The \emph{Fermi} density discussed in sub-section \ref{Perturbative nets} is an example of a higher-hierarchy density, built on top of the \emph{Boltzmann} density, subject to additional constraints  \cite{Landau80},  section 53. 
Clusterization aggregates the low-hierarchy partial equilibria - the observations -  into higher-hierarchy partial equilibria - clusters, sub-section \ref{Probabilistic and}.

To mimic this universal phenomenon, in its second, generative, installment, ACE  combines a classifier and a generative auto-encoder in the same space of observables,   in a brand of an auto-encoder  \emph{supervision}, Figure \ref{Fig1.4}. ACE generalizes the classic idea of using separate decoders for separate classes, \cite{Hinton95}.  In training, the conditional latent density $p(\mathbf{z} |\mathbf{x}_{\mu})$ from sub-section \ref{Probabilistic and} is generalized to $p(\mathbf{z} |\mathbf{x}_{\mu}, c_{\mu})$, where $c_{\mu}$ is the class or label of the $\mu$-th observation. Since, of course, classification labels can not be used on the testing set, the sampling during testing is from a mixture of densities, with class probabilities supplied by the classifier (hence the dashed line in Figure \ref{Fig1.4}). Mixture densities in the posterior were also used in \cite{Kingma14-3}, albeit in a different architecture. The ACE is universal in the sense of subsection \ref{Universality} and achieves state-of-the-art performance, both as a classifier and a density estimator, Figure \ref{Fig4.3}. For its relation with information geometry, see open problem \ref{How is} in section \ref{Open problems}.
 
\section{Theoretical background.}
\label{Theoretical background}
\subsection{Definitions.}
\label{Conditional densities}
For discrete data, our minimization target - the cross-entropy (\ref{1.-1}) - can be decomposed as the sum:
 \begin{align}
-\log \mathcal{L}(r||q)=\mathcal{S}(r) + \mathcal{D}(r||q),
 \label{2.-3}
 \end{align}
of the \emph{Kullback-Leibler divergence} $\mathcal{D}(r||q)$ between \emph{empirical} $r()$ and model $q()$ densities:
 \begin{align}
\mathcal{D}(r||q) :=\mathbb{E}_{r(\mathbf{x})}[ \log r(\mathbf{x}) - \log q(\mathbf{x})],
\label{2.-2}
\end{align}
and the standard \emph{ Boltzmann entropy} $\mathcal{S}(r)$ of $r(.)$:
\begin{align}
\mathcal{S}(r) :=\mathbb{E}_{r(\mathbf{x})}[-\log r(\mathbf{x})],
\label{2.-1}
\end{align}
 $\mathcal{S}(.)$ $\neq \mathcal{S}^E(.)$. Because the entropy $\mathcal{S}(r)$ of $r(.)$ does not depend on our model distribution $q(.)$, the minimization of the cross-entropy is equivalent to minimizing the Kullback-Leibler divergence $\mathcal{D}(r||q)$. It is common in statistical physics, see e.g.  \cite{Naudts10}, to formally consider $-\mathcal{D}(r||q)$  as a \emph{generalized entropy} $\mathcal{S}_q(r)$, for the case of a non-trivial base measure $q()$  - see (\ref{2.23d}) below.

As discussed in sub-section \ref{Probabilistic and}, the latents $\{\mathbf{z}\}$  in generative nets are sampled  from a closed-form conditional model  density $p(\mathbf{z}|\mathbf{x})$.  The latents are of course not given a priori, so the joint empirical density is:
\begin{align}
r(\mathbf{x,z}) =\frac{1}{P}\sum_{\mu} \delta(\mathbf{x} - \mathbf{x}_{\mu})p(\mathbf{z}|\mathbf{x}),
\label{2.0}
\end{align}
with corresponding marginal empirical densities $r(\mathbf{x})$ $=\frac{1}{P}\sum_{\mu} \delta(\mathbf{x}- \mathbf{x}_{\mu})$, $r(\mathbf{z})$ $=\frac{1}{P}\sum_{\mu} p(\mathbf{z}| \mathbf{x}_{\mu})$ (\cite{Kulhavy96}, section 2.3; \cite{Cover06}, problem 3.12). Hence, the cross-entropy of $q(\mathbf{x})$ $:=\int p(\mathbf{x,z})d\mathbf{z}$ $= p(\mathbf{x,z})/p(\mathbf{z|x})$ is an arithmetic average\footnote{This decomposition does not imply independence of observations: the latent variables can in general contain information from more than one observation, as for example in the case of time series auto-regression.} across observations $-\mathcal{L}(r||q)$ $= -\frac{1}{P}\sum_{\mu} \log q(\mathbf{x}_{\mu})$. From the Bayes identity, we have for our optimization target (\ref{1.-1}), in terms of the joint density:
\begin{align}
&-\log \mathcal{L}(r||q) := \mathbb{E}_{r(\mathbf{x})}[-\log q(\mathbf{x})] = \nonumber \\
&=\mathbb{E}_{r(\mathbf{x,z})}[-\log p(\mathbf{x,z})]   + \mathbb{E}_{r(\mathbf{x,z})}[\log p(\mathbf{z}|\mathbf{x})].
\label{2.1}
\end{align}
From the explicit form (\ref{2.0}) of $r(\mathbf{x,z})$, for the $\mu$-th observation:
\begin{align}
- \log q(\mathbf{x}_{\mu}) =\mathcal{D}( p(\mathbf{z| x_{\mu}}) || p(\mathbf{x_{\mu},z})) =\nonumber \\
=\mathbb{E}_{p(\mathbf{z}|\mathbf{x}_{\mu})}[-\log p(\mathbf{x_{\mu},z})]   -  \mathcal{S} (p(\mathbf{z}|\mathbf{x}_{\mu})) ,
\label{2.2}
\end{align}
where $\mathcal{S} (p(\mathbf{z}|\mathbf{x}_{\mu}))$ $ =\mathbb{E}_{p(\mathbf{z | x}_{\mu})}[-\log p(\mathbf{z | x}_{\mu})] $ is the Boltzmann entropy of the  model distribution, conditional on a given observation $\mathbf{x}_{\mu}$. If we sample the latent observables only once per observation, as commonly done, the right-hand side reduces to $-  \log p(\mathbf{x_{\mu},z_{\mu}})$ $+  \log p(\mathbf{z_{\mu}|x_{\mu}})$.

\subsection{Conditional independence.}
\label{Conditional independence}
The hidden/latent observables $\mathbf{z}$ $= \{\mathbf{z}_j\}_{j=1}^{N_{lat}}$ are \emph{conditionally independent} if, for a given observation $\mathbf{x}_{\mu}$, one has $p(\mathbf{z} | \mathbf{x}_{\mu})$ $ = \prod_{j=1}^{N_{lat}} p(\mathbf{z}_j | \mathbf{x}_{\mu})$. From the independence bound of the Boltzmann entropy $\mathcal{S}(p(\mathbf{z}|\mathbf{x}_{\mu}))$ $ \leq \sum_{j=1}^{N_{lat}} \mathcal{S}( p(\mathbf{z}_{j}|\mathbf{x}_{\mu}))$, \cite{Cover06}, Chapter 2, conditional independence  minimizes the negative entropy term on the right-hand side of (\ref{2.2}). Everything else being equal, conditional independence is hence optimal  for nets. 

\subsection{Exponential/Gibbs class of densities.}
\label{Gibbs and}

There is a broad class of probability density families - \emph{Gibbs} a.k.a. \emph {canonical} or \emph{exponential} families - which dominate the choices of model densities, both in physics and neural nets. This class includes a sufficiently large number of density families: Gaussian, Bernoulli, exponential, gamma, etc. Their general closed form is:
\begin{align}
p_{\boldsymbol{\lambda}}(\mathbf{z}) = \frac{p(\mathbf{z})}{\mathcal{Z}} e^{-\sum_{s=1}^M \lambda_{s} \mathcal{M}_{s}(\mathbf{z}) },
\label{2.11}
\end{align}
where $p(\mathbf{z})$ is an arbitrary \emph{base} or \emph{prior} density, $\boldsymbol{\lambda}$ $ = \{\lambda_s \}$ are  \emph{Lagrange } multipliers a.k.a. \emph{natural parameters}, $\mathcal{M}_{j}(\mathbf{z})$ are so-called \emph{sufficient statistics}, and  $\mathcal{Z}$ $=\mathcal{Z}(\boldsymbol{\lambda})$ is the normalizing \emph{partition function}. Knowing $\mathcal{Z}$ is equivalent to knowing the \emph{free energy} $\mathcal{F}(\boldsymbol{\lambda})$ $= -\log \mathcal{Z}(\boldsymbol{\lambda})$, which allows to re-write (\ref{2.11}) as:
\begin{align}
p_{\boldsymbol{\lambda}}(\mathbf{z}) = p(\mathbf{z})e^{\mathcal{F} -\sum_{s=1}^M \lambda_{s} \mathcal{M}_{s}(\mathbf{z}) } = p(\mathbf{z})e^{\mathcal{F}(\boldsymbol{\lambda}) -\boldsymbol{\lambda}. \boldsymbol{\mathcal{M}(\mathbf{z})} },
\label{2.21}
\end{align}
 where $\boldsymbol{\lambda} . \boldsymbol{\mathcal{M}}$ is the scalar product of the vectors $\boldsymbol \lambda$ $=\{\lambda_s\}$ and $\boldsymbol {\mathcal{M}}$ $=\{\mathcal{M}_s\}$.

\subsection{Macroscopic quantities.}
\label{Macroscopic quantities}

In physics, the expectations of the sufficient statistics $\mathbf{m}$ $=\mathbb{E}_{p_{\lambda}(\mathbf{z})}[\boldsymbol{\mathcal{M}}(\mathbf{z})]$  form a complete set of \emph{macroscopic} a.k.a. \emph{thermodynamic quantities} or state variables like energy, momenta, number of particles, etc, fully describing the ${\mu}$-th conditional  state,  sub-section \ref{Probabilistic and}, see   \cite{Landau80}, sections 28,34,35,110. In neural nets, the sufficient statistics  are typically  monomials like $\mathcal{M}_{1}(\mathbf{z}) = \mathbf{z}$, $\mathcal{M}_{2}(\mathbf{z}) = \mathbf{z}^2$, etc, whose  expectations form a vector of  moments. As proposed in sub-section \ref{Symmetries in}, one can add them to the list the symmetry statistics, see section \ref{Latent symmetry} for details. From the definition of free energy, one derives immediately that it is a generative function of the expectations $\mathbf{m}$:
\begin{align}
\frac{\partial \mathcal{F}(\boldsymbol{\lambda})}{\partial \boldsymbol{\lambda}} =  \mathbf{m}(\boldsymbol{\lambda}),
\label{2.21a}
\end{align}
and, more generally, of their higher n-th moments:
\begin{align}
\frac{\partial \mathcal{F}(\boldsymbol{\lambda})}{\partial \lambda_{i}\partial \lambda_{j}...\partial \lambda_{k}} = (-1)^{n+1} \mathbb{E}_{p_{\boldsymbol{\lambda}}(\mathbf{z})}[\mathcal{M}_{i}(\mathbf{z})\mathcal{M}_{j}(\mathbf{z})...\mathcal{M}_{k}(\mathbf{z})].
\label{2.21b}
\end{align}

\subsection{Variational Pythagorean theorem.}
\label{Variational Pythagorean}

The exponential/Gibbs class of families is special because it is the variational \emph{maximum entropy} class: when the base density $p()$ is trivial, it is the unique functional form which maximizes the Boltzmann entropy $\mathcal{S}(f)$ across the universe $\{f(\mathbf{z}) \}$ of all densities  with given macroscopic quantities $\mathbf{m}$ $=\mathbb{E}_{f}[\boldsymbol{\mathcal{M}}(\mathbf{z})]$, see \cite{Cover06}, chapter 12. The natural parameters $\boldsymbol{\lambda}$ $ = \boldsymbol{\lambda}(\mathbf{m})$  are computed so as to satisfy these constraints.  They are Lagrange constraints multipliers in the variational calculus derivation of the maximum entropy property.

\begin{figure}[!ht]
\begin{tikzpicture}
\shade[left color=gray!10,right color=gray!40] 
  (0,-2) to[out=-10,in=150] (5,-2) -- (8,-0.5) to[out=150,in=-10] (3.5,-0.5) -- cycle;
  
\node[fill,circle, label=above:$f(\mathbf{z})$, inner sep=0pt, minimum size=0.pt, draw] at (6,0.5) (a) {x};  
\node[fill,circle,  label=below:$p_{\lambda}(\mathbf{z})$, inner sep=0.pt, minimum size=0.pt, draw] at (6,-1) (b) {y};  
\node[fill,circle,   label=below:$p(\mathbf{z})$, inner sep=0.pt, minimum size=0.pt, draw] at (2,-1.2) (c) {y}; 

\path [->](a) edge   node [xshift = 0.in] {$\mathcal{D}\left( f(\mathbf{z}) || p_{\lambda}(\mathbf{z}) \right)$}  (b);
\path [->, dashed](b) edge [bend right=20] node [yshift = -0.2in] {$\mathcal{D}\left( p_{\lambda}(\mathbf{z}) || p(\mathbf{z}) \right)$}  (c);
\path [->](a) edge node [xshift = -0.5in] {$\mathcal{D}\left( f(\mathbf{z}) || p(\mathbf{z}) \right)$}   (c); 
\end{tikzpicture}
\caption{A naive visualization of the probabilistic (variational) Pythagorean theorem from (\ref{2.12}).}
\label{Fig2.3}
\end{figure}
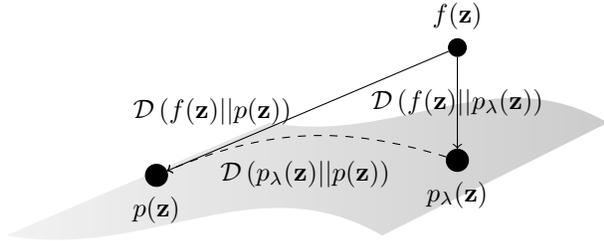

The Gibbs class is special in even stronger sense: it is a \emph{minimum divergence} class. For an arbitrary base density $p(\mathbf{z})$, the Kullback-Leibler divergence  $\mathcal{D} (p_{\boldsymbol{\lambda}}(\mathbf{z}) || p(\mathbf{z}))) $ minimizes the divergence $\mathcal{D} (f(\mathbf{z}) || p(\mathbf{z})) $ across the  universe $\{f(\mathbf{z}) \}$ of all densities with given macroscopic quantities $\mathbf{m}$ $=\mathbb{E}_{f}[\boldsymbol{\mathcal{M}}(\mathbf{z})]$. This follows  from the \emph{variational Pythagorean theorem}, Figure \ref{Fig2.3}, \cite{Chentsov68}, \cite{Kulhavy96}, section 3.3:
\begin{align}
\mathcal{D}(f(\mathbf{z}) || p(\mathbf{z}))& = \mathcal{D} (f(\mathbf{z}) || p_{\boldsymbol{\lambda}}(\mathbf{z})) + \mathcal{D} (p_{\boldsymbol{\lambda}} (\mathbf{z})|| p(\mathbf{z})) \geq  \nonumber \\
& \geq \mathcal{D} (p_{\boldsymbol{\lambda}} (\mathbf{z})|| p(\mathbf{z})). 
\label{2.12}
\end{align}

\subsection{Generative error.}
\label{Generative error}
As we will see in sub-section \ref{Cross-entropy decomposition}, minimizing the divergence $\mathcal{D} (f(\mathbf{z}) || p(\mathbf{z})) $ across an unknown a priori family of  conditional distributions $p(\mathbf{z}| \mathbf{x}_{\mu})$ $=f(\mathbf{z})$, is crucial for the quality of a generative net.  The smaller this divergence, the more likely are the sampled from  $p(\mathbf{z})$ newly created objects to resemble the training set. The minimum divergence property (\ref{2.12}) implies that we are always better off choosing $p(\mathbf{z}| \mathbf{x}_{\mu})$ from a Gibbs class, as in (\ref{2.13}), hence the name Gibbs machines. We will refer to the minimum divergence $\mathcal{D}( p(\mathbf{z| x}_{\mu}) || p(\mathbf{z}))$ as \emph{generative error}:  
\begin{align}
\mathcal{D}^{gen}(\mathbf{m}(\mathbf{x}_{\mu})) &:= \mathcal{D}( p(\mathbf{z| x}_{\mu}) || p(\mathbf{z})) \geq 0.
\label{2.12a}
\end{align}
The lack of explicit dependence on the natural parameters $\boldsymbol{\lambda}$ on the left  is because $\mathcal{D}^{gen}$ depends on them only indirectly, via the macroscopic quantities $\mathbf{m}$ $=\mathbf{m}(\boldsymbol{\lambda})$ (see sub-section \ref{Generative thermodynamic}).

\subsection{Conditional latent densities.}
\label{Conditional latent}
For a given  observation $\mathbf{x}_{\mu}$, choosing a conditional latent density from the Gibbs type (\ref{2.21}) is equivalent to:
\begin{align}
p(\mathbf{z}| \mathbf{x}_{\mu}) := p(\mathbf{z}) e^{\mathcal{F}^{gen}(\mathbf{x}_{\mu}, \boldsymbol{\lambda})- \boldsymbol{ \lambda}. \boldsymbol{\mathcal{M}}(\mathbf{z}) },
\label{2.13}
\end{align}
where the superscript `gen' is short for \emph{generative}. We will see shortly that the free energy $\mathcal{F}^{gen}$ and the negative generative error $-\mathcal{D}^{gen}$ from the previous sub-section are \emph{concave conjugates}, hence the shared superscript. In practice, in order for $p(\mathbf{z}| \mathbf{x}_{\mu})$ to be tractable, $p(\mathbf{z})$ has to be from a specific parametric family within the Gibbs class: then both $p(\mathbf{z})$ and $p(\mathbf{z}| \mathbf{x}_{\mu})$ will be tractable and in the same family, e.g. Gaussian, exponential, etc. 

Except for the symmetry statistics introduced in sections \ref{Symmetries in}, \ref{Latent symmetry}, the macroscopic quantities $\mathbf{m}(\mathbf{x}_{\mu})$   $=\mathbb{E}_{p(\mathbf{z}| \mathbf{x}_{\mu})}[\boldsymbol{\mathcal{M}}(\mathbf{z})]$ for the $\mu$-th quantum state are free parameters. Together with the the symmetry statistics, they can be thought of as \emph{quantum numbers} distinguishing the observations, a.k.a. partial equilibrium states, from one another, in the spirit of quantum statistics  \cite{Landau80}, section 5, see sub-section \ref{Probabilistic and} here. 
These quantum numbers are added to the rest of the free net parameters, to be optimized by standard methods, like back-propagation/ stochastic gradient descent, etc.

\subsection{Boltzmann-Gibbs thermodynamic identity.}
\label{Boltzmann-Gibbs thermodynamic}

The identities relating various macroscopic (thermodynamic) quantities are referred to in statistical physics as \emph{thermodynamic identities}. Recall that the classic Boltzmann-Gibbs distribution 
is a special case of the exponential/Gibbs distribution (\ref{2.21}), with a trivial base density $p()$, and only one sufficient statistics - the \emph{microscopic energy } or \emph{Hamiltonian} $\mathcal{H}(,)$:
\begin{align}
p^{BG}(\mathbf{z|x_{\mu}})=  e^{\mathcal{F}(\mathbf{x}_{\mu}) - \beta \mathcal{H}(\mathbf{x_{\mu},z})},
\label{2.22}
\end{align}
where  $\beta$ $:= \frac{1}{T}$ is the inverse \emph{temperature} and the only natural parameter. Taking logs and expectations w.r.t. $p^{BG}(\mathbf{z|x_{\mu}})$, one gets the classic thermodynamic identity for the free energy $\mathcal{F}^{BG}$ $=\mathcal{F}^{BG}(\mathbf{x}_{\mu}, \beta)$ and entropy $\mathcal{S}\big( p^{BG}(\mathbf{z| x}_{\mu})\big)$:
\begin{align}
\mathcal{F}^{BG}(\beta) =  \beta \mathcal{U}^{BG}(\beta) - \mathcal{S}(\mathcal{U}^{BG}(\beta)),
\label{2.23}
\end{align}
where  the only macroscopic quantity - the energy $\mathcal{U}^{BG}$ - is the expectation $\mathcal{U}^{BG}$ $:=\mathbb{E}_{p^{BG}(\mathbf{z}| \mathbf{x}_{\mu})}[\mathcal{H}(\mathbf{x_{\mu},z})]$. The entropy $\mathcal{S}$ depends on $\beta$ only indirectly, via the energy i.e. $\mathcal{S}$ $=\mathcal{S}(\mathcal{U}^{BG}(\beta))$. 

The increase of free energy  has the physics interpretation of work needed to be done by the outside environment, in order to increase the macroscopic quantity  of the system, i.e. the energy in this case. The negative sign in (\ref{2.23}) confirms the intuition that, for the same increase of internal energy $\mathcal{U}^{BG}$, less ordered systems need less work from the outside.

\subsection{Generative thermodynamic identity.}
\label{Generative thermodynamic}

As mentioned in sub-section \ref{Conditional densities}, in a more general context like (\ref{2.13}), the negative generative error $-\mathcal{D}^{gen}( \mathbf{m}) )$ plays the role of generalized entropy $\mathcal{S}_p(p(\mathbf{z| x}_{\mu}))$, with base measure $p(\mathbf{z})$. As is standard in statistical mechanics, it can be shown (\cite{Kulhavy96}, Chapter 3) that $-\mathcal{D}^{gen}( \mathbf{m}) )$ is a concave  function, expressable via the Legendre transform  of its conjugate - the generative free energy $\mathcal{F}^{gen}(\boldsymbol {\lambda})$:
\begin{align}
-\mathcal{D}^{gen}(\mathbf{m}) =\min_{\boldsymbol {\lambda}} \{\boldsymbol {\lambda} .\mathbf{m} - \mathcal{F}^{gen}(\boldsymbol {\lambda})  \},
\label{2.23a}
\end{align}
or, equivalently,
\begin{align}
\mathcal{D}^{gen}(\mathbf{m}) =\max_{\boldsymbol {\lambda}} \{\boldsymbol {-\lambda} .\mathbf{m} + \mathcal{F}^{gen}(\boldsymbol {\lambda})  \},
\label{2.23a1}
\end{align}
where $\boldsymbol {\lambda} .\mathbf{m}$ is a scalar product of the vectors $\boldsymbol {\lambda}$ and $\mathbf{m}$ and $\boldsymbol {\lambda}$ is allowed to run free. Note that, while the variational Pythagorean theorem establishes the minimum property of $\mathcal{D}^{gen}$ as a \emph{functional} $\mathcal{D}^{gen}(f||p)$ over the space of functions $\{f(\mathbf{z})\}$, $\mathcal{D}^{gen}$ is a maximum when viewed as  an explicit \emph{function} $\mathcal{D}^{gen}(\mathbf{m},\boldsymbol {\lambda})$ of both the macroscopic quantities $\mathbf{m}$ and the natural parameters $\boldsymbol {\lambda}$. Equivalently,
\begin{align}
\mathcal{F}^{gen}(\boldsymbol {\lambda}) =\min_{\mathbf m} \{\boldsymbol {\lambda} .\mathbf{m} + \mathcal{D}^{gen}(\mathbf{m})  \},
\label{2.23b}
\end{align} 
where $\mathbf m$ is allowed to run freely. At the optimal point (``equilibrium'') $\mathbf{m}^{gen}$, this implies the dual identity of  (\ref{2.21a}):
\begin{align}
-\frac{\partial \mathcal{D}^{gen}(\mathbf m)}{\partial \mathbf m} \Bigg{|}_{\mathbf{m} = \mathbf{m}^{gen}} =:  \boldsymbol{\lambda}^{gen}(\mathbf{m}^{gen}).
\label{2.23c}
\end{align}  
Similarly to (\ref{2.21a}), the derivative on the left-hand side  defines the function $\boldsymbol{\lambda}^{gen}$  $=\boldsymbol{\lambda}^{gen}(\mathbf m)$ as a function of $\mathbf m$ everywhere. While generative free energy was defined in (\ref{2.23b}) for every  $\boldsymbol{ \lambda}$, unless the function in (\ref{2.23c}) is invertible, its image  is only a subset of the full space of natural parameters $\boldsymbol{ \lambda}$. Assuming such invertibility,  the generalization of (\ref{2.23}) is: 
\begin{align}
\mathcal{F}^{gen}(\boldsymbol{\lambda}) = \boldsymbol{ \lambda} . \mathbf{m}^{gen}(\boldsymbol{ \lambda}) + \mathcal{D}^{gen}(\mathbf m^{gen}(\boldsymbol{ \lambda})),
\label{2.23d}
\end{align}
where we skipped for brevity the dependence on $\mathbf{x}_{\mu}$. Counter to classic thermodynamics (\ref{2.23}), we have a plus sign? The two factors comprising the divergence $\mathcal{D}^{gen}(||)$, clearly work against each other. On the one hand, the negative entropy term $ -\mathcal{S}( p(\mathbf{z| x}_{\mu}))$ $\leq 0$ reduces free energy as usual: less ordered systems take less work to create. On the other hand, when the system resides in an infinitely-large ``thermostat'' of non-trivial density $p(\mathbf{z})$, the cross-entropy term $\mathbb{E}_{p(\mathbf{z|x}_{\mu})}[-\log p(\mathbf{z})]$ $\geq 0$, increases the free energy back up. It  measures the amount of work it takes to counter the thermostat's influence.

The chain rule and (\ref{2.21a}), (\ref{2.23c}),   give in addition the indirect dependence of $\mathcal{D}^{gen}$ on the natural parameters $\boldsymbol{ \lambda}$:
\begin{align}
\frac{\partial \mathcal{D}^{gen}(\mathbf {m}(\boldsymbol{ \lambda}))}{\partial \boldsymbol{ \lambda}}  =  -\boldsymbol{\lambda} . \frac{\partial \mathbf{m}(\boldsymbol{ \lambda})}{\partial \boldsymbol{ \lambda}} =
-\boldsymbol{\lambda} . \frac{\partial^{2} \mathcal{F}^{gen}(\boldsymbol{ \lambda})}{\partial \boldsymbol{ \lambda}\partial \boldsymbol{ \lambda}}.
\label{2.24}
\end{align} 
From (\ref{2.21b}), this derivative is conveniently expressed in terms of the matrix of second moments of the sufficient statistics:
\begin{align}
\frac{\partial \mathcal{D}^{gen}(\mathbf {m}(\boldsymbol{ \lambda}))}{\partial \boldsymbol{ \lambda}}  = 
\boldsymbol{\lambda} . \mathbb{E}_{p(\mathbf{z|x}_{\mu})}[\boldsymbol{\mathcal{M}}(\mathbf{z}) \boldsymbol{\mathcal{M}}(\mathbf{z}) ].
\label{2.25}
\end{align}

\subsection{q-Gibbs densities.}
\label{q-Gibbs densities}
Gibbs densities are only a special case (for $q=1$) of the broad class of  q-Gibbs (or q-exponential) densities.  The corresponding \emph{nonextensive} statistical mechanics \cite{Tsallis09}, describes more adequately long-range-interacting many-body systems like  typical human-generated data sets. By virtue of  replacing the exponential with a q-exponential, $\log$ with q-$\log$ and defining a respective q-entropy, most of the formalism of classic thermodynamics has been generalized. Many of the properties of the exponential class remain true for the q-exponential class \cite{Naudts10}, \cite{Amari11}, see open problem \ref{q-Gibbs} in section \ref{Open problems}.

\section{Application to generative neural nets.}
A \emph {fully-generative net} creates original observations by sampling from an unconditional model density  $p(\mathbf{z})$, unencumbered by  any observations $\{\mathbf{x}_{\mu}\}$. \emph{Perturbative nets} on the other hand, do not contain an  unconditional model density $p(\mathbf{z})$. They rely instead on an initial observation $\mathbf{x}_{\mu}$, a conditional density $p(\mathbf{z|x}_{\mu})$, and the decomposition (\ref{2.2}) for parameter estimation.
\subsection{Perturbative nets.}
\label{Perturbative nets}

The most successful family of perturbative nets to date have been the  \emph{Boltzmann machines} \cite{Smolensky86} 
and their multiple re-incarnations.  Assuming completeness of the state variables, Boltzmann machines adopt as joint model density the Boltzmann density, a special case of the Boltzmann-Gibbs equilibrium density from sub-section \ref{Gibbs and}: $p^{BG}(\mathbf{x_{\mu},z})$ $ = \frac{1}{\mathcal{Z}(\mathbf{x}_{\mu})} e^{-\frac{1}{T}\mathcal{H}(\mathbf{x_{\mu},z})}$, with a single sufficient statistics - a  bi-linear Hamiltonian function $\mathcal{H}(,)$, a trivial base density and temperature $T=1$. It  is not tractable because the \emph{partition function} $\mathcal{Z}(\mathbf{x}_{\mu})$ can not be computed in closed form. On the other hand, for discrete data, the conditional density of the \emph{restricted} Boltzmann machines (RBM) is  tractable and is the familiar  Fermi density  $p^{F} (\mathbf{z|x_{\mu}}) = 1/(1+ e ^{\mathcal{H}(\mathbf{x_{\mu},z})} )$. The intractable joint density term in (\ref{2.2}) is handled by approximations of its gradients  like \emph{contrastive divergence}  \cite{Hinton02}, to avoid brute-force  Monte Carlo methods averaging impossibly many paths. 

Due to the simple bi-linear shape of the Hamiltonian, RBM-s have  latent variables conditionally independent on the visible variables (and vice versa). 
Despite their limitations  when handling non-binary data, deep Boltzmann machines \cite{Sal09} 
have until recently been the dominant  universal nets: They perform well both as classifiers \cite{Srivastava14} and probability density estimators \cite{Sal08}.

\subsection{Reconstruction error.}
\label{Reconstruction error}
A neural net is said to have \emph{reconstruction} capabilities, if it has a \emph{decoder}, which for a given latent $\mathbf{z}$, can assign a \emph{reconstruction density} $p^{rec}(\mathbf{x_{\mu}|z})$ of an observation $\mathbf{x}_{\mu}$. Following standard procedure, the \emph{reconstruction error} is the usual cross-entropy (\ref{1.-1}) of this  reconstruction density and  the empirical density (\ref{2.0}):
\begin{align}
 -\log\mathcal{L}^{rec}(\mathbf{x}_{\mu}) := \mathbb{E}_{ p(\mathbf{z}|\mathbf{x}_{\mu})}[- \log p^{rec}(\mathbf{x}_{\mu}| \mathbf{z})] \geq 0.
 \label{3.-1}
 \end{align}
Reconstruction densities are typically from the exponential/Gibbs class - Bernoulli, Gaussian, etc -  with unity covariance matrix, and a trivial base density $p()$. But the key reconstruction macroscopic quantity - the expectation $\mathbf{ m}^{rec}(\mathbf{x}_{\mu})$ $=\mathbb{E}_{ p(\mathbf{z}|\mathbf{x}_{\mu})}[\mathbf {x}_{\mu}]$ - is generally an intractable function of $\mathbf{z}$, given by the net decoder. 

The cross-entropy $- \log\mathcal{L}^{rec}$ depends on the generative natural parameters only via the expectation density $p(\mathbf{z}|\mathbf{x}_{\mu})$, and its derivatives are from (\ref{3.-1}):
\begin{align}
-\frac{\partial \log\mathcal{L}^{rec}(\mathbf {m}(\boldsymbol{ \lambda}))}{\partial \boldsymbol{ \lambda}}  =   \mathbb{E}_{p(\mathbf{z|x}_{\mu})}[\boldsymbol{\mathcal{M}}(\mathbf{z}) \log p^{rec}(\mathbf{x}_{\mu}| \mathbf{z})].
\label{3.1}
\end{align}

\subsection{Fully-generative nets. Regimes.}
\label{Fully-generative nets}
 We will distinguish two separate regimes of operation of a fully-generative net: 
\begin{enumerate}[nolistsep]
\item
\textbf{non-creative regime}: This is the common regime for net training, validation or testing. Latent observables   are sampled from a closed-form model conditional density $p(\mathbf{z|x})$, as in sub-section \ref{Conditional latent}, with observations $\{\mathbf{x}_{\mu}\}$ attached to the net.   A closed-form reconstruction model density $p^{rec}(\mathbf{x|z})$, as in sub-section \ref{Reconstruction error}, is also chosen. 
\item
\textbf{creative regime}: The net has been trained and latent observables  $\mathbf{z}$ $= \{\mathbf{z}_j\}_{j=1}^{N_{lat}}$ are sampled from a closed-form model density $p(\mathbf{z})$, unencumbered by  observations $\{\mathbf{x}_{\mu}\}$. The same reconstruction  density $p^{rec}(\mathbf{x|z})$  as in the non-creative regime is used.
\end{enumerate}

The joint density is chosen to be:
\begin{align}
p(\mathbf{x,z}) :=p(\mathbf{z})p^{rec}(\mathbf{x|z}).
\label{3.1a}
\end{align}
 The empirical densities are the same as in sub-section \ref{Conditional densities}.

\subsection{Variational error.}
\label{Variational error}

The implied marginal density  $q(\mathbf{x})$ $:=\int p(\mathbf{x,z})d\mathbf{z}$ and the implied conditional density $q(\mathbf{z|x})$ $=p(\mathbf{x,z})/q(\mathbf{x})$  are generally intractable in fully generative nets. Moreover, the implied conditional $q(\mathbf{z|x})$ is of course different than our chosen \emph{a priori} $p(\mathbf{z|x})$ in the non-creative regime. For a given observation $\mathbf{x}_{\mu}$, the divergence between the two is called \emph{variational error}:
\begin{align}
\mathcal{D}^{var}(\mathbf{x}_{\mu}) = \mathcal{D}( p(\mathbf{z| x}_{\mu}) || q(\mathbf{z| x}_{\mu}) ) \geq 0.
\label{3.2}
\end{align}
Its optimization is the subject of the so-called \emph{fixed-form} variational Bayes analysis \cite{Saul95}. We will discuss its estimation in sub-section \ref{Estimating variational}, see also open problem \ref{Variational Bayes}, section \ref{Open problems}. 

\subsection{Cross-entropy decomposition.}
\label{Cross-entropy decomposition}
Following standard procedure, our training minimization target is the cross-entropy (\ref{1.-1}) between the  marginal density $q(\mathbf{x})$ $=\int p(\mathbf{x,z})d\mathbf{z}$ and the  non-creative empirical density (\ref{2.0}). Expanding the  joint density (\ref{3.1a}) in both Bayesian directions, one can decompose in terms of the implied conditional $ q(\mathbf{z| x}_{\mu})$ as follows:
\begin{align}
&-\log \mathcal{L}(r||q) := \mathbb{E}_{r(\mathbf{z})}[-\log p(\mathbf{z})]  + \nonumber \\
&+\mathbb{E}_{r(\mathbf{x,z})}[- \log p^{rec}(\mathbf{x}|\mathbf{z})]
+ \mathbb{E}_{r(\mathbf{x,z})}[ \log q(\mathbf{z}|\mathbf{x})].
\label{3.3}
\end{align}
From the explicit form (\ref{2.0}) of $r(\mathbf{x,z})$, for the $\mu$-th observation $\mathbf{x}_{\mu}$:
\begin{align}
&-\log q(\mathbf{x}_{\mu}) = \mathbb{E}_{p(\mathbf{z}|\mathbf{x}_{\mu})} [-\log p(\mathbf{z})] +\nonumber \\
&+ \mathbb{E}_{ p(\mathbf{z}|\mathbf{x}_{\mu})} [- \log p^{rec}(\mathbf{x}_{\mu}| \mathbf{z})]
+ \mathbb{E}_{p(\mathbf{z}|\mathbf{x}_{\mu})}[\log q(\mathbf{z}|\mathbf{x}_{\mu})].
\label{3.4}
\end{align}
Subtracting $\mathcal{S}(p(\mathbf{z}|\mathbf{x}_{\mu}))$ from the first term and adding it to the third, and using the definitions  (\ref{2.12a}) of generative error $\mathcal{D}^{gen}$, (\ref{3.-1})  of reconstruction error $-\log\mathcal{L}^{rec}$  and  (\ref{3.2}) of variational error $\mathcal{D}^{var}$, the final expression for our minimization target is:
\begin{align}
-\log q(\mathbf{x}_{\mu}) =  \mathcal{D}^{gen}(\mathbf{x}_{\mu}) -\log\mathcal{L}^{rec}(\mathbf{x}_{\mu}) - \mathcal{D}^{var}(\mathbf{x}_{\mu}).
\label{3.5}
\end{align}
Let us highlight the essence and computability of each of the three components:
\begin{enumerate}[nolistsep]
 \item
$\mathcal{D}^{gen} \geq 0$: the generative error (\ref{2.12a}) is the  divergence between the generative densities in the non-creative and creative regimes. Minimizing it ensures the general similarity of objects generated in the two regimes. It  can be interpreted as the hypotenuse in the variational Pythagorean theorem (\ref{2.12}) and is computable in closed form for many Gibbs/exponential densities. 
\item
 $-\log\mathcal{L}^{rec} \geq 0$: the reconstruction error (\ref{3.-1}) measures the negative likelihood of getting $\mathbf{x}_{\mu}$ back, after the transformations and  randomness inside the net.  It can be computed by any net endowed with a decoder via standard Monte Carlo averaging, as traditional \emph{auto-encoders} do. Importantly for training, in order to compute gradients with respect to the generative macroscopic quantities $\mathbf{m}^{gen}$, a change of variable is needed, replacing sampling from $p(\mathbf{z}|\mathbf{x}_{\mu})$ by sampling from $p(\mathbf{z})$.  Such transformations exists for many of the exponential/Gibbs probability families \cite{Kingma14-1}.

\item
$\mathcal{D}^{var} \geq 0$: the variational error (\ref{3.2}) measures the divergence between our chosen functional form $p(\mathbf{z| x}_{\mu})$ for latent density and the \emph{implied} by the net latent density $q(\mathbf{z| x}_{\mu})$. Due to the intractability of  $q(\mathbf{z|x_{\mu}})$, the variational error has to be computed numerically via Monte Carlo methods, see sub-section \ref{Estimating variational} and open problem \ref{Variational Bayes}, section \ref{Open problems}.
\end{enumerate}

\subsection{Cross-entropy upper bound.}
\label{Upper bound}
Dropping the variational error in (\ref{3.5}) yields an upper bound $\mathcal{B}()$ for the cross-entropy\footnote{This is an expanded version of the textbook  \emph{variational inequality}  \cite{Cover06}, Exercise 8.6.}  $-\log q(\mathbf{x}_{\mu})$: 
\begin{align}
 \mathcal{B}(\mathbf{x}_{\mu} ) &:= 
  \mathcal{D}^{gen}(\mathbf{x}_{\mu}) - \log\mathcal{L}^{rec}(\mathbf{x}_{\mu}) = \nonumber \\
&=\mathcal{D}( p(\mathbf{z}|\mathbf{x}_{\mu}) || p(\mathbf{x}_{\mu},\mathbf{z})   ).
\label{3.6}
\end{align} 
While the last  expression is formally equivalent to the general expression (\ref{2.2}) for $-\log q(\mathbf{x}_{\mu})$, the density  $p(\mathbf{z|x}_{\mu})$ in (\ref{2.2})  is the correct conditional density of the joint density  $p(\mathbf{x}_{\mu},\mathbf{z})$, while $p(\mathbf{z|x}_{\mu})$ here is merely an approximation to the implied  conditional  $q(\mathbf{z|x}_{\mu})$.

From (\ref{2.25}) and (\ref{3.1}), the derivative of the upper bound with respect to the generative natural parameters is:
\begin{align}
\frac{\partial \mathcal{B}(\boldsymbol{ \lambda})}{\partial \boldsymbol{ \lambda}}  =   \boldsymbol{\lambda} . \mathbb{E}_{p(\mathbf{z|x}_{\mu})}[\boldsymbol{\mathcal{M}}(\mathbf{z}) \boldsymbol{\mathcal{M} }(\mathbf{z}) ] + \nonumber \\
+\mathbb{E}_{p(\mathbf{z|x}_{\mu})}[ \log p^{rec}(\mathbf{x}_{\mu}| \mathbf{z})\boldsymbol{\mathcal{M}}(\mathbf{z})].
\label{3.6a}
\end{align}
At optima or inflection points of $\mathcal{B}(\boldsymbol{ \lambda})$, the derivative is zero and, if the moment matrix $\mathbb{E}_{p(\mathbf{z|x}_{\mu})}[\boldsymbol{\mathcal{M}}(\mathbf{z}) \boldsymbol{\mathcal{M} }(\mathbf{z}) ]$ is invertible, the generative natural parameters become:
\begin{align}
\boldsymbol{\lambda}  = - \mathbb{E}_{p(\mathbf{z|x}_{\mu})}[ \log p^{rec}(\mathbf{x}_{\mu}| \mathbf{z})\boldsymbol{\mathcal{M}}(\mathbf{z})] . \nonumber \\
\mathbb{E}_{p(\mathbf{z|x}_{\mu})}[\boldsymbol{\mathcal{M}}(\mathbf{z}) \boldsymbol{\mathcal{M} }(\mathbf{z}) ]^{-1}.
\label{3.6b}
\end{align}

\subsection{Gibbs machines.}
\label{Gibbs machines}
It is clear from the above derivation  that (\ref{3.5}), (\ref{3.6}) are universal for fully-generative nets and were hence used in the first fully-generative nets, the VAE-s \cite{Kingma14-1}, \cite{Rezende14}. The VAE-s owe their name to the variational error term and were introduced in the context of very general sampling densities. Everything else being equal, the variational Pythagorean theorem (\ref{2.12}) implies that latent sampling densities (\ref{2.13}) from the Gibbs class minimize the generative error. Hence we call the respective nets Gibbs machines. While the variational error is due to an approximation, the variational principle from which the Gibbs class is derived, is fundamental to  statistical mechanics. 

\subsection{Estimating variational error.}
\label{Estimating variational}

A closer look at the equilibrium identity for the natural parameters (\ref{3.6b}) reveals that it is identical to the maximum likelihood estimates of the coefficients of a linear regression:
\begin{align}
 \log p^{rec}(\mathbf{x}_{\mu}| \mathbf{z}) = \alpha - \boldsymbol{\lambda} . \boldsymbol{\mathcal{M}}(\mathbf{z}) + \epsilon(\mathbf{x}_{\mu}, \mathbf{z}), 
\label{3.7}
\end{align}
where $\alpha$ is an intercept and $\epsilon$ an error term. This observation was first made in \cite{Richard07} and later used in \cite{Salimans13} to estimate variational error in the context of Variational Bayes. Adding $\log p(\mathbf{z})$ to both sides, adding/subtracting $\mathcal{F}^{gen}$ to the right side, and recalling (\ref{2.13}), (\ref{3.1a}), transforms (\ref{3.7}) into:
\begin{align}
 \log p(\mathbf{x}_{\mu}, \mathbf{z}) = \alpha - \mathcal{F}^{gen} + \log p( \mathbf{z}|\mathbf{x}_{\mu})  + \epsilon(\mathbf{x}_{\mu}, \mathbf{z}). 
\label{3.8}
\end{align}
Requiring $\mathbb{E}_{p( \mathbf{z}|\mathbf{x}_{\mu})}[\epsilon(\mathbf{x}_{\mu}, \mathbf{z})]$ $=0$ yields for the regression intercept the last expression in (\ref{3.6}), with a negative sign:
\begin{align}
 \log p(\mathbf{x}_{\mu}, \mathbf{z}) =  -\mathcal{B}(\mathbf{x}_{\mu} ) + \log p( \mathbf{z}|\mathbf{x}_{\mu})  + \epsilon(\mathbf{x}_{\mu}, \mathbf{z}). 
\label{3.9}
\end{align} 
The implied marginal density $q(\mathbf{x}_{\mu})$ $:=\int p(\mathbf{x_{\mu},z})d\mathbf{z}$ is now:
\begin{align}
 q(\mathbf{x}_{\mu}) =  e^{-\mathcal{B}(\mathbf{x}_{\mu} )} \mathbb{E}_{ p( \mathbf{z}|\mathbf{x}_{\mu})}[ e^{\epsilon(\mathbf{x}_{\mu}, \mathbf{z})}], 
\label{3.10}
\end{align} 
hence the cross-entropy for the $\mu$-th observation is:
\begin{align}
 - \log q(\mathbf{x}_{\mu}) =  \mathcal{B}(\mathbf{x}_{\mu} ) - \log \mathbb{E}_{ p( \mathbf{z}|\mathbf{x}_{\mu})}[ e^{\epsilon(\mathbf{x}_{\mu}, \mathbf{z})}], 
\label{3.11}
\end{align} 
i.e.
\begin{align}
 \mathcal{D}^{var}(\mathbf{x}_{\mu}) =  \log \mathbb{E}_{ p( \mathbf{z}|\mathbf{x}_{\mu})}[ e^{\epsilon(\mathbf{x}_{\mu}, \mathbf{z})}]. 
\label{3.12}
 \end{align} 
This can be estimated either via Monte Carlo methods, or in a closed form. Assuming for example that $\epsilon(\mathbf{x}_{\mu}, \mathbf{z})$ $\sim \mathcal{N}(0,\sigma(\mathbf{x}_{\mu})^2)$ is a Gaussian with variance $\sigma(\mathbf{x}_{\mu})^2$, yields $\mathcal{D}^{var}(\mathbf{x}_{\mu})$ $= \frac{\sigma(\mathbf{x}_{\mu})^2}{2}$.

While not rigorous, it is interesting to use these  $\mathcal{D}^{var}(\mathbf{x}_{\mu})$ estimates, to train the neural net with the full cross-entropy  $-q(\mathbf{x}_{\mu})$ from (\ref{3.5}), and not just its upper bound $\mathcal{B}(\mathbf{x}_{\mu})$ from (\ref{3.6}) - see open problem \ref{Variational Bayes}, section \ref{Open problems}.

\subsection{Generative conditional independence.}
\label{Generative conditional}
Without offering a rigorous proof, we believe that the conditional independence argument from sub-section \ref{Conditional independence} can be generalized to this context: For  a given reconstruction error, the generative error $\mathcal{D}^{gen}$ and hence the upper bound  (\ref{3.6}) is minimized when the latent variables are conditionally independent. For Gaussian multi-variate sampling, this follows  from the explicit form of the generative error \cite{Gil13}, table 3, and \emph{Hadamard's} inequality \cite{Cover06}, chapter 8.

\section{Latent symmetry statistics. Momenta.}
\label{Latent symmetry}
We will show for brevity only how to build spatial invariances  in a 2-dim square visual recognition model. For real-life data sets, the symmetry statistics have to be computed via another net, see \cite{Georgiev15-2}.

Every observable i.e. pixel $\mathbf{x}_i$ $, i=0,...,N$, can be assigned  a horizontal $h_i$ and a vertical  $v_i$ integer coordinates on the screen,  e.g., $h_i,v_i$  $\in \{1,...,\sqrt{N}\}$.  In these coordinates, a row-observation $\mathbf{x}_{\mu}$  $=\{x_{\mu i} \}_{i=1}^N$ becomes a matrix-observation $\{x_{\mu, h_i,v_i} \}$ and a net layer of size $N$ becomes a layer of size $\sqrt{N} \times \sqrt{N}$. 
The  \emph{center of mass} $(h_{\mu}, v_{\mu})$:
\begin{align}
 h_{\mu} := \frac{\sum_{i=1}^N x_{\mu i} h_i}{ \sum_i x_{\mu i}},\quad v_{\mu} := \frac{\sum_{i=1}^N x_{\mu i} v_i}{ \sum_i x_{\mu i}}.
 \label{3.11} 
\end{align}
of every observation defines latent symmetry statistics   $\mathbf{h}$ $=\{ h_{\mu} \}_{\mu=1}^P$ and $\mathbf{v}$ $=\{ v_{\mu} \}_{\mu=1}^P$, see  sub-sections \ref{Symmetries in}, \ref{Gibbs and}. Without loss of generality, we assumed here that $x_{\mu i} \geq 0$, hence $0 \leq$ $ h_{\mu},v_{\mu}$ $ \leq \sqrt{N}$. In the coordinate system centered by $(h_{\mu}, v_{\mu})$, we have for every $\mu$ the new coordinates:
\begin{align}
 (\hat{h}_{\mu i},\hat{v}_{\mu i}) := ( h_i - h_{\mu}, v_i - v_{\mu}),
\label{3.12}
\end{align}
$-\sqrt{N} \leq$ $\hat{h}_{\mu i},\hat{v}_{\mu i}$   $\leq \sqrt{N}$. In this coordinate system, every pixel has polar coordinates $r_{\mu i}$ $:=\sqrt{ \hat{h}_{\mu i}^2 +  \hat{v}_{\mu i}^2 }$, $\varphi_{\mu i}$ $:= atan2 (\hat{v}_{\mu i},\hat{h}_{\mu i} )$ $\in (-\pi, \pi]$, hence the new symmetry statistics: \emph{scale} $\mathbf{r}$ $=\{ r_{\mu} \}_{\mu=1}^P$ and \emph{angle} $\boldsymbol{\varphi}$ $=\{ \varphi_{\mu} \}_{\mu=1}^P$: 
\begin{align}
 r_{\mu} := \frac{\sum_{i=1}^N x_{\mu i} r_{\mu i}}{ \sum_i x_{\mu i}},\quad \varphi_{\mu} := \frac{\sum_{i=1}^N x_{\mu i} \varphi_{\mu i}}{ \sum_i x_{\mu i}},
\label{3.14}
\end{align}
$0 < r_{\mu} \leq \sqrt{2N}$, $-\pi < \varphi_{\mu} \leq \pi$. In order to un-scale and un-rotate an image, change coordinates one more time to:
\begin{align}
(\tilde{h}_{\mu i}, \tilde{v}_{\mu i} ) := \frac{C}{r_{\mu}}(\hat{h}_{\mu i}, \hat{v}_{\mu i} ) \left( \begin{matrix} \cos(-\varphi_{\mu}) &\sin(-\varphi_{\mu}) \\ 
\sin(\varphi_{\mu}) &\cos(-\varphi_{\mu})\end{matrix}\right) ,
\label{3.15}
\end{align}
$-M \leq$ $\tilde{h}_{\mu i},\tilde{v}_{\mu i}$   $\leq M$, for some constants $C, M$ depending\footnote{The scale $r_{\mu}$ typically needs  to have a lower bound, in order to ensure that $M$ is of the same order of magnitude as $\sqrt{N}$.}  on $\min\limits_{\mu} r_{\mu}$, $\max\limits_{\mu} r_{\mu}$. After rounding and a shift, we thus have for any observation $\mu$,  an  index mapping:
\vskip -0.2in
\begin{align}
\{1,...,N\} & \rightarrow \{1,...,(2M+1)^2\} \nonumber \\
i & \rightarrow \tilde{h}_{\mu i} + (\tilde{v}_{\mu i}-1) (2M+1).
\label{3.16}
\end{align}
When $2M+1$ $>\sqrt{N}$, the $(2M+1 - \sqrt{N})^2$ indexes which are not in the mapping image,  correspond to identically zero observables for that observation. In the  coordinates (\ref{3.15}), a layer of size $N$ becomes a layer of size $(2M+1)^2$. 

In summary: i) for auto-encoders, apply mapping (\ref{3.16}) at the input and its inverse at the output of the net; ii) for classifiers,  apply mapping (\ref{3.16}) at the input only; iii) for both, include in addition the  symmetry statistics $\mathbf{h,v,r},\boldsymbol{\varphi}$  in the latent layer, if needed, see \cite{Georgiev15-2}. The prior  model density $p()$ of symmetry statistics can be assumed equal to their parametrized posterior $p(.|\mathbf{x}_{\mu})$, but there are other options.

When sampling the symmetry statistics from independent Laplacians as in (\ref{1.2}) e.g., the respective density means are set to be $h_{\mu}, v_{\mu}, r_{\mu}, \varphi_{\mu}$ from (\ref{3.11}), (\ref{3.14}). The  density scales $\sigma^{h}_{\mu}, \sigma^v_{\mu}, \sigma^r_{\mu}, \sigma^{\varphi}_{\mu}$ on the other hand are free parameters, and can  in principle be optimized in the non-creative regime, alongside the rest of the net parameters, sub-section \ref{Conditional latent}. As argued in sub-section \ref{Symmetries in},  the inverted scales are  the  scaled \emph{momenta}. In the creative regime, when sampling e.g. from $\mathbf{h}$ alone, one will get horizontally shifted identical replicas. See open problem \ref{Invariants}, section \ref{Open problems}.

\section{Experimental results.}
\label{Experimental results}
The Theano  \cite{Theano12} code used for  experiments is in \cite{Georgiev15-1}, see also \cite{Popov15}.
\subsection{Non-generative ACE.}
\label{Non-generative ACE}

The motivation for the non-generative ACE comes from the Einstein observation entropies $ \{- \log p^G(\mathbf{x_{\mu}}) \}_{\mu}$, as in sub-section \ref{Observation entropies}, and their relation to singular value decomposition (SVD). Recall that  the SVD of  the $B \times N$ data matrix $\mathbf{X}$ with $B$ observations and $N$ observables is $\mathbf{X}$ $=\mathbf{V}\boldsymbol{\Lambda}\mathbf{W}^T$. The $B \times B$ matrix $\mathbf{VV}^T$ is i) a projection mapping; ii) its diagonals are up to a  constant the  negative Einstein observation entropies from sub-section (\ref{Observation entropies}), i.e., the Gaussian log-likelihoods $ \{- \log p^G(\mathbf{x_{\mu}}) \}_{\mu}$; and iii) it is invariant on $\mathbf{X}$, i.e. $\mathbf{X}$ $=\mathbf{VV}^T \mathbf{X}$.
\begin{figure*}[!ht]
\begin{minipage}[]{\columnwidth}
\begin{tikzpicture}
\node[rectangle, text width =0.6 \columnwidth,  text  centered,  draw ] 
	(Input) at (5,0) {\small \underline{ 1. \textbf{Input layer}}: \\$\mathbf{x}_{\mu}$ \\Size = \# observables $N$ } ;
\node[rectangle,  text width = 0.5 \columnwidth,  text  centered, below=0.2inof Input, draw] 
	(Latent/Feature)  {\small \underline{2. \textbf{Latent hidden layer}}: \\ $\mathbf{h}_{\mu}=\phi(\mathbf{x}_{\mu} \mathbf{W}^{(1)} + \mathbf{b}^{(1)}) $ \\Size = $N_{lat}$};
\node[rectangle,  text width = 0.6 \columnwidth,  text  centered, below=0.2inof Latent/Feature, draw] 
	(Output)  {\small \underline{3. \textbf{Output layer}}: \\ $\mathbf{\hat{x}_{\mu}}=\varphi(\mathbf{h}_{\mu} \mathbf{W}^{(2)} + \mathbf{b}^{(2)}) $ 
	\\Size = $N$};
\path [->](Input) edge node {} (Latent/Feature); 
\path [->](Latent/Feature) edge node {} (Output); 
\draw [->]  ([xshift=0.2in]Output.east)  edge [bend right=0, thick, double]   node [rotate=270, midway, yshift = 0.2in] { Back-propagation of $\min(-\log\mathcal{L}_{recon})$} ([xshift=0.2in]Input.east);
\draw [decorate, decoration={brace}]  ([xshift=-0.35in, yshift=0.2in]Latent/Feature.west) -- ([xshift=-0.2in]Input.west) node [midway,rotate=90,yshift=0.2in] {Encoder};
\draw [decorate, decoration={brace}]  ([xshift=-0.2in]Output.west) -- ([xshift=-0.35in, yshift=-0.2in]Latent/Feature.west) node [midway,rotate=90,yshift=0.2in] {Decoder};
\end{tikzpicture} 
\end{minipage}\hfill
\begin{minipage}[]{\columnwidth}
\begin{tikzpicture}
\node[rectangle, text width = 0.6 \columnwidth, text  centered,  draw ] 
	(Input) at (5,0) {\small \underline{1. \textbf{Input layer}}: \\ $\mathbf{x}_{i}$ \\Size = \# observations $P$} ;
\node[rectangle,  text width = 0.5 \columnwidth, text  centered, below=0.2inof Input, draw] 
	(Hidden)  {\small \underline{2. \textbf{Latent hidden layer}}: \\ $\mathbf{h'}_{i}=\phi(\mathbf{V}^{(1)} \mathbf{x}_{i} + \mathbf{b}^{(1)}) $ \\Size = $N_{lat}$};
\node[rectangle,  text width = 0.6 \columnwidth, text  centered, below=0.2inof Hidden, draw] 
	(Output)  {\small \underline{3. \textbf{Output layer}}: \\ $\mathbf{\hat{x}}_{i} = \varphi(\mathbf{V}^{(2)}\mathbf{h'}_{i} + \mathbf{b}^{(2)}) $ \\
	Size = $P$};
\path [->](Input) edge node {} (Latent/Feature); 
\path [->](Latent/Feature) edge node {} (Output);
\draw [->]  ([xshift=0.2in]Output.east)   edge [bend right=0, thick, double]  node [rotate=270, midway,  yshift = 0.2in] { Back-propagation of $\min(-\log\mathcal{L}'_{recon})$} ([xshift=0.2in]Input.east);
\end{tikzpicture}
\end{minipage}\hfill
\caption{Shallow auto-encoder in the space of observables (left) and observations (right). Minimization targets are the reconstruction errors in the respective spaces $-\log\mathcal{L}_{recon}$ and $-\log\mathcal{L}'_{recon}$ defined for binarized data in Appendix \ref{Software and}, $\phi$, $\varphi$ are non-linearities.}
\label{Fig4.1}
\end{figure*}
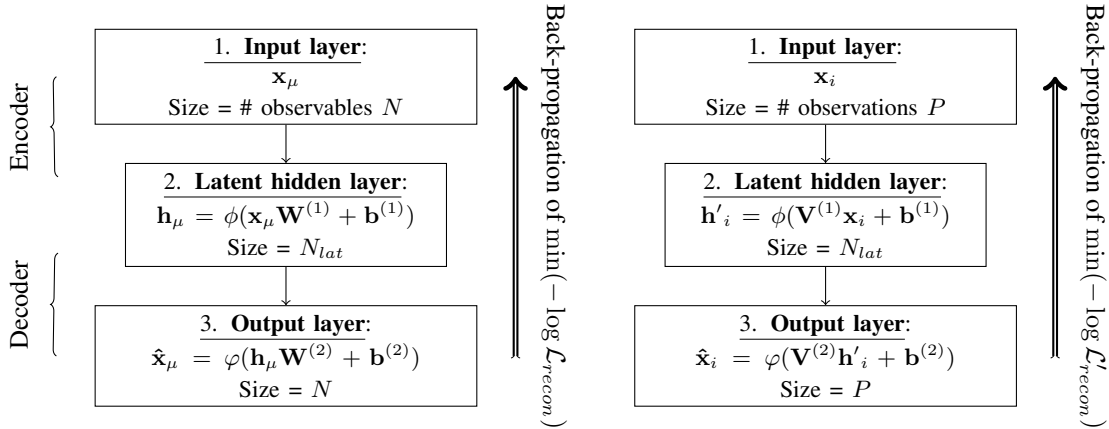

Let us consider a shallow auto-encoder, Figure \ref{Fig4.1}, left, and its dual in the space of observations,  Figure \ref{Fig4.1}, right. It can be shown that for tied weights $\mathbf{V}^{(2)}$ $=\mathbf{V}^{(1)T}$, in the absence of non-linearities, the optimal  $B \times N_{lat}$ hidden layer solution $\mathbf{H}_{o}$ on the left is   $\mathbf{H}_{o}$ $=\mathbf{V}^{(2)}$  $=\mathbf{V}$ \cite{Georgiev15-3}. The divergence of $\mathbf{VV}^T \mathbf{X}$ from $\mathbf{X}$ is thus the reconstruction error in the dual space and minimizing it gets us closer to the optimal $\mathbf{H}_o$. If we treat the first  hidden layer $\mathbf{H}$ of a classifier as the rescaled dual weight matrix $\sqrt{B}\mathbf{V}^{(2)}$,  we arrive at the dual reconstruction error :
\begin{align}
-\log \mathcal{L}'_{recon}=\frac{1}{B}\sum_{i=1}^N \mathbb{E}_{\mathbf{x}_{i}}[- \log \varphi(\mathbf{H H}^T\mathbf{x}_{i}/B)],
\label{4.1}
\end{align}
for a given column-vector observable $\mathbf{x}_i=$ $\{x_{\mu i}\}_{\mu=1}^B$ and sigmoid non-linearity $\varphi()$, see Appendix \ref{Software and}.

The non-generative ACE has  as  minimization target the composite cross-entropy (\ref{1.1}), with $-\log\mathcal{L}_{AE}$ replaced by the  dual reconstruction error (\ref{4.1}). The orthogonality $\mathbf{V}^T\mathbf{V}$ $=\mathbf{I}_{N_{lat}}$ of $\mathbf{V}$ implies the need for an additional \emph{batch normalization}, similar to \cite{Ioffe15}, see Appendix \ref{Software and}.

\begin{figure}[!ht]
\begin{minipage}{0.5\columnwidth}
\includegraphics[width=\columnwidth]{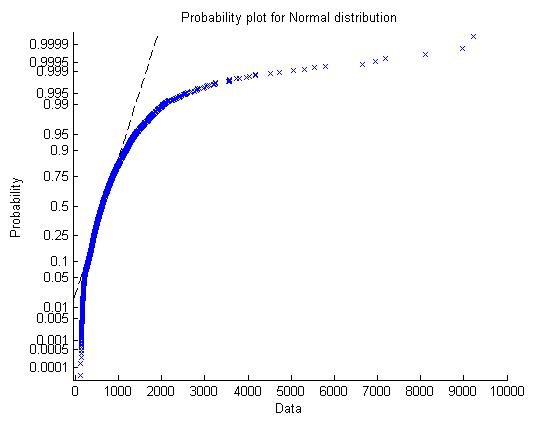}
\end{minipage}%
\begin{minipage}{\columnwidth}
\includegraphics[width=0.5\columnwidth]{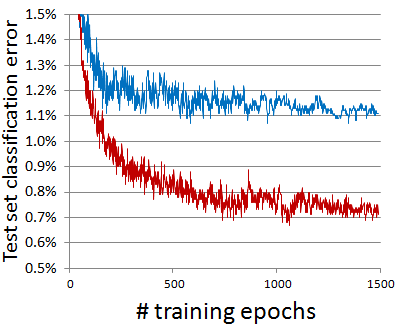}
\end{minipage}  \hfill
\caption{ \textbf{Left.} The same Q-Q plot as on the right of Figure \ref{Fig1.1}, but for the non-generative ACE, sub-section \ref{Non-generative ACE}, with the same hyper-parameters as on the right of of Figure \ref{Fig1.1}.  
\textbf{Right.} Classification error for the MNIST 10000 test set, as a function of training \emph{epochs}, i.e., one full swipe over all training observations. The top line is the standard classifier as on the right of Figure \ref{Fig1.1}. The bottom line is the classification error of the non-generative ACE  with the same hyper-parameters. }
\label{Fig4.2}
\vskip -0.2in
\end{figure}

The best known results for the test classification error of feed-forward, non-convolutional nets, without artificial data  augmentation, are in the 0.9-1\% handle \cite{Srivastava14}, table 2. As shown on the right of Figure \ref{Fig4.2}, the non-generative ACE offers a 20-30\%  improvement.

\subsection{Generative ACE.}
\label{Generative ACE}

\begin{figure}[!ht]
\begin{minipage}{0.5\columnwidth}
\includegraphics[width=\columnwidth]{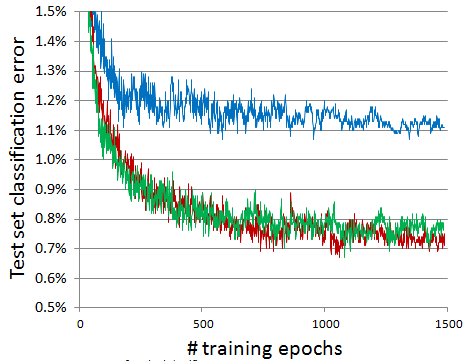}
\end{minipage}%
\begin{minipage}{\columnwidth}
\includegraphics[width=0.5\columnwidth]{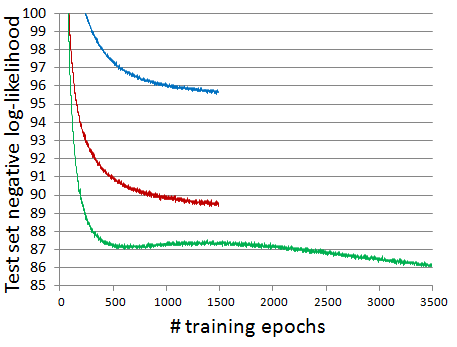}
\end{minipage}  \hfill
\caption{ \textbf{Left.} Classification error for the MNIST 10000 test set. Top line is from a standard classifier net as on the right of Figure \ref{Fig1.1}. Bottom lines are from generative ACE in classification mode: Gaussian sampling (red) and Laplacian (green). Layer sizes 784-700-(100x10)-(700x10)-(784x10) for the AE branch and 784-700-700-700-10 for the C branch, Figure \ref{Fig1.4} and Appendix \ref{Software and}, learning rate = 0.0015, decay = 500 epochs, batch size = 10000. The dual reconstruction error $-\log\mathcal{L}'_{recon}$ from sub-section \ref{Non-generative ACE} is added to the overall cost.  
\textbf{Right.} Upper bound (\ref{3.6})  of the negative log-likelihood for the binarized MNIST 10000 test set. The top line is the standard Gibbs machine with Gaussian sampling, layer sizes 784-700-400-700-784 and other hyper-parameters as below. The middle line is the same net but with Laplacian sampling. The bottom line is generative ACE with Laplacian mixture sampling. Layer sizes 784-700-(400x10)-(700x10)-(784x10) for the AE branch and 784-700-700-700-10 for the C branch, Figure \ref{Fig1.4} and Appendix \ref{Software and}, learning rate = 0.0002, decay = 500 epochs, batch size = 1000. }
\label{Fig4.3}
\vskip -0.2in
\end{figure}

The architecture is in Figure \ref{Fig1.4}, the minimization target is the ACE cross-entropy (\ref{1.1}), with $-\log\mathcal{L}_{AE}$ replaced by the upper bound (\ref{3.6}). Laplacian sampling density  is used in training  and the mixed Laplacian in testing, with the explicit formulas for the generative error in Appendix \ref{Laplacian and}. The generative ACE produces similarly outstanding classification results as the non-generative ACE on the regular MNIST data set,  Figure \ref{Fig4.3}, left. Even without tweaking hyper-parameters, it also produces outstanding results for the density estimation of the binarized MNIST data set, Figure \ref{Fig4.3}, right. An upper bound for the  negative log-likelihood in the 86-87 handle is in the ballpark of the best non-recurrent nets, \cite{Gregor15}, table 2.

\section{Open problems.} 
\label{Open problems} 
\begin{enumerate}[nolistsep]

\item
Use the freely available intricates, sub-section \ref{The curse}, directly as feature detectors, in lieu of artificially computed Independent Component Analysis (ICA) features \cite{Hyvarinen09}.
\label{Intricates}

\item
Test empirically the performance of ACE, with the symmetry statistics added as in section \ref{Latent symmetry},  computed either in closed form, or from specialized nets, as in \cite{Jadeberg15}. For the distorted MNIST and CIFAR10 datasets, see  \cite{Georgiev15-2}.
\label{Invariants}

\item
Deepen and make generative the shallow dual encoder of the non-generative ACE, sub-sections \ref{ACE}, \ref{Non-generative ACE} here.
\label{Generative}

\item
Test empirically q-Gibbs machines, with cross-entropies replaced by their q-equivalents and sampling from q-Gibbs densities, sub-section \ref{q-Gibbs densities} here.
\label{q-Gibbs}

\item
Improve the upper bound for the generative error of mixture densities in Appendix \ref{Laplacian and}, by using variational methods as in \cite{Hershey07}.
\label{Use a}

\item
How is the ACE blend of exponential and mixture densities related to the beautiful duality between these two families, underlying information geometry \cite{Amari00}, section 3.7?
\label{How is}

\item
Estimate the variational error (\ref{3.2}) \cite{Richard07}, \cite{Salimans13} and use it in training to minimize the full cross-entropy, not merely its  upper bound, as suggested in sub-section \ref{Estimating variational} here.
\label{Variational Bayes}
\end{enumerate}

\section*{Acknowledgments} 
We appreciate motivating discussions with Ivaylo Popov and Nikola Toshev. Credit goes to Christine Haas for coining the terms \emph{intricates} and \emph{creative/non-creative} regimes. 

\bibliographystyle{abbrvnat} 
\bibliography{bibliographyNN}

\begin{thebibliography}{41}
\providecommand{\natexlab}[1]{#1}
\providecommand{\url}[1]{\texttt{#1}}
\expandafter\ifx\csname urlstyle\endcsname\relax
  \providecommand{\doi}[1]{doi: #1}\else
  \providecommand{\doi}{doi: \begingroup \urlstyle{rm}\Url}\fi

\bibitem[Amari and Nagaoka(2001)]{Amari00}
S.~Amari and H.~Nagaoka.
\newblock \emph{Methods of Information Geometry}.
\newblock American Mathematical Society, 2001.

\bibitem[Amari and Ohara(2011)]{Amari11}
S.-i. Amari and A.~Ohara.
\newblock Geometry of q-exponential family of probability distributions.
\newblock \emph{Entropy}, 13\penalty0 (6):\penalty0 1170--1185, 2011.

\bibitem[Bastien et~al.(2012)Bastien, Lamblin, Pascanu, Bergstra, Goodfellow,
  Bergeron, Bouchard, and Bengio]{Theano12}
F.~Bastien, P.~Lamblin, R.~Pascanu, J.~Bergstra, I.~J. Goodfellow, A.~Bergeron,
  N.~Bouchard, and Y.~Bengio.
\newblock Theano: new features and speed improvements, 2012.

\bibitem[Chentsov(1968)]{Chentsov68}
N.~Chentsov.
\newblock Nonsymmetrical distance between probability distributions, entropy
  and the theorem of {P}ythagoras.
\newblock \emph{Mathematical notes of the Academy of Sciences of the USSR},
  4\penalty0 (3):\penalty0 686--691, 1968.

\bibitem[Coates et~al.(2011)Coates, Lee, and Ng]{Coates11}
A.~Coates, H.~Lee, and A.~Y. Ng.
\newblock An analysis of single-layer networks in unsupervised feature
  learning.
\newblock In \emph{AISTATS}, 2011.

\bibitem[Cover and Thomas(2006)]{Cover06}
T.~M. Cover and J.~A. Thomas.
\newblock \emph{Elements of Information Theory, 2nd Edition}.
\newblock Wiley-Interscience, 2006.

\bibitem[Einstein(2006)]{Einstein10}
A.~Einstein.
\newblock On {Boltzmann’s} principle and some immediate consequences thereof.
\newblock In T.~Damour, O.~Darrigol, B.~Duplantier, and V.~Rivasseau, editors,
  \emph{Einstein, 1905 2005}, volume~47 of \emph{Progress in Mathematical
  Physics}, pages 183--199. Birkhäuser Basel, 2006.

\bibitem[Gelfand and Fomin(1963)]{Gelfand63}
I.~Gelfand and S.~Fomin.
\newblock \emph{Calculus of Variations}.
\newblock Prentice-Hall, 1963.

\bibitem[Georgiev(2015{\natexlab{a}})]{Georgiev15-1}
G.~Georgiev.
\newblock {ACE}, 2015{\natexlab{a}}.
\newblock https://github.com/galinngeorgiev/ACE.

\bibitem[Georgiev(2015{\natexlab{b}})]{Georgiev15-2}
G.~Georgiev.
\newblock Symmetries and control in generative neural nets.,
  2015{\natexlab{b}}.
\newblock arXiv:1511.02841.

\bibitem[Georgiev(2015{\natexlab{c}})]{Georgiev15-3}
G.~Georgiev.
\newblock Duality between observables and observations in neural nets.,
  2015{\natexlab{c}}.
\newblock to appear.

\bibitem[Gil et~al.(2013)Gil, Alajaji, and Linder]{Gil13}
M.~Gil, F.~Alajaji, and T.~Linder.
\newblock R{\'e}nyi divergence measures for commonly used univariate continuous
  distributions.
\newblock \emph{Information Sciences}, 249:\penalty0 124–--131, 2013.

\bibitem[Goodfellow et~al.(2013)Goodfellow, Warde-Farley, Mirza, Courville, and
  Bengio]{Goodfellow13}
I.~J. Goodfellow, D.~Warde-Farley, M.~Mirza, A.~Courville, and Y.~Bengio.
\newblock Maxout networks, 2013.
\newblock arXiv:1302.4389.

\bibitem[Gregor et~al.(2015)Gregor, Danihelka, Graves, and Wierstra]{Gregor15}
K.~Gregor, I.~Danihelka, A.~Graves, and D.~Wierstra.
\newblock {DRAW:} {A} recurrent neural network for image generation, 2015.
\newblock arXiv:1502.04623.

\bibitem[Hershey and Olsen(2007)]{Hershey07}
J.~Hershey and P.~Olsen.
\newblock Approximating the {Kullback Leibler} divergence between {Gaussian}
  mixture models.
\newblock In \emph{ICASSP}, volume~4, pages 317--320, 2007.

\bibitem[Hinton(2002)]{Hinton02}
G.~E. Hinton.
\newblock Training products of experts by minimizing contrastive divergence.
\newblock \emph{Neural Computation}, 14:\penalty0 1771--1800, 2002.

\bibitem[Hinton et~al.(1995)Hinton, Revow, and Dayan]{Hinton95}
G.~E. Hinton, M.~Revow, and P.~Dayan.
\newblock Recognizing handwritten digits using mixtures of linear models.
\newblock In \emph{AINIPS}, volume~7, pages 1015--1022, 1995.

\bibitem[Hyvarinen et~al.(2009)Hyvarinen, Hurri, and Hoyer]{Hyvarinen09}
A.~Hyvarinen, J.~Hurri, and P.~O. Hoyer.
\newblock \emph{Natural Image Statistics: A probabilistic approach to early
  computational vision}.
\newblock Springer-Verlag, 2009.

\bibitem[Ioffe and Szegedy(2015)]{Ioffe15}
S.~Ioffe and C.~Szegedy.
\newblock Batch normalization: Accelerating deep network training by reducing
  internal covariate shift, 2015.
\newblock arXiv:1502.03167.

\bibitem[Jadeberg et~al.(2015)Jadeberg, Symonyan, Zisserman, and
  Kavukcuoglu]{Jadeberg15}
M.~Jadeberg, K.~Symonyan, A.~Zisserman, and K.~Kavukcuoglu.
\newblock Spatial transformer networks, 2015.
\newblock arXiv:1412.6980.

\bibitem[Kingma and Ba(2015)]{Kingma14-2}
D.~Kingma and J.~Ba.
\newblock Adam: A method for stochastic optimization.
\newblock In \emph{ICLR}, 2015.

\bibitem[Kingma and Welling(2014)]{Kingma14-1}
D.~P. Kingma and M.~Welling.
\newblock Auto-encoding variational {B}ayes.
\newblock In \emph{ICLR}, 2014.

\bibitem[Kingma et~al.(2014)Kingma, Rezende, Mohamed, and Welling]{Kingma14-3}
D.~P. Kingma, D.~J. Rezende, S.~Mohamed, and M.~Welling.
\newblock Semi-supervised learning with deep generative models.
\newblock 2014.
\newblock arXiv:1406.5298.

\bibitem[Kulhav{\`y}(1996)]{Kulhavy96}
R.~Kulhav{\`y}.
\newblock \emph{Recursive Nonlinear Estimation: A Geometric Approach}.
\newblock Lecture Notes in Control And Information Sciences. Springer-Verlag
  GmbH, 1996.

\bibitem[Landau and Lifshitz(1977)]{Landau77}
L.~Landau and E.~Lifshitz.
\newblock \emph{Quantim Mechanics, Non-relativistic theory, 3rd edition}.
\newblock Pergamon Press, 1977.

\bibitem[Landau and Lifshitz(1980)]{Landau80}
L.~Landau and E.~Lifshitz.
\newblock \emph{Statistical Physics, Part 1, 3rd edition}.
\newblock Elsevier Science, 1980.

\bibitem[Le et~al.(2011)Le, Karpenko, Ngiam, and Ng]{Le11}
Q.~V. Le, A.~Karpenko, J.~Ngiam, and A.~Y. Ng.
\newblock {ICA} with reconstruction cost for efficient overcomplete feature
  learning.
\newblock In \emph{NIPS}, volume~24, pages 1017--1025, 2011.

\bibitem[LeCun et~al.(1998)LeCun, Cortes, and Burges]{LeCun98}
Y.~LeCun, C.~Cortes, and C.~J. Burges.
\newblock {MNIST} handwritten digit database, 1998.
\newblock http://yann.lecun.com/exdb/mnist/.

\bibitem[Mardia et~al.(1979)Mardia, Kent, and Bibby]{Mardia79}
K.~V. Mardia, J.~T. Kent, and J.~M. Bibby.
\newblock \emph{{Multivariate Analysis}}.
\newblock Academic Press, 1979.

\bibitem[Naudts(2010)]{Naudts10}
J.~Naudts.
\newblock The q -exponential family in statistical physics.
\newblock \emph{Journal of Physics: Conference Series}, 201\penalty0
  (1):\penalty0 012003, 2010.

\bibitem[Popov(2015)]{Popov15}
I.~Popov.
\newblock Theano-{L}ights, 2015.
\newblock https://github.com/Ivaylo-Popov/Theano-Lights.

\bibitem[Rezende et~al.(2014)Rezende, Mohamed, and Wierstra]{Rezende14}
D.~J. Rezende, S.~Mohamed, and D.~Wierstra.
\newblock Stochastic backpropagation and approximate inference in deep
  generative models.
\newblock In \emph{JMLR}, volume~32, 2014.

\bibitem[Richard and Zhang(2007)]{Richard07}
J.-F. Richard and W.~Zhang.
\newblock Efficient high-dimensional importance sampling.
\newblock \emph{Journal of Econometrics}, 141\penalty0 (2), 2007.

\bibitem[Rifai et~al.(2012)Rifai, Bengio, Dauphin, and Vincent]{Rifai12}
S.~Rifai, Y.~Bengio, Y.~Dauphin, and P.~Vincent.
\newblock A generative process for sampling contractive auto-encoders.
\newblock In \emph{ICML}, 2012.

\bibitem[Salakhutdinov and Hinton(2009)]{Sal09}
R.~Salakhutdinov and G.~Hinton.
\newblock Deep {B}oltzmann machines.
\newblock In \emph{AISTATS}, volume~5, pages 448--455, 2009.

\bibitem[Salakhutdinov and Murray(2008)]{Sal08}
R.~Salakhutdinov and I.~Murray.
\newblock On the quantitative analysis of deep belief networks.
\newblock In \emph{ICML}, volume~25, 2008.

\bibitem[Salimans and Knowles(2013)]{Salimans13}
T.~Salimans and D.~A. Knowles.
\newblock Fixed-form variational posterior approximation through stochastic
  linear regression.
\newblock \emph{Bayesian Analysis}, 8:\penalty0 837--882, 2013.

\bibitem[Saul and Jordan(1995)]{Saul95}
L.~Saul and M.~I. Jordan.
\newblock Exploiting tractable substructures in intractable networks.
\newblock In \emph{NIPS}, volume~8, 1995.

\bibitem[Smolensky(1986)]{Smolensky86}
P.~Smolensky.
\newblock Information processing in dynamical systems: Foundations of harmony
  theory.
\newblock In D.~E. Rumelhart and J.~L. McClelland, editors, \emph{Parallel
  Distributed Processing}, volume~1, chapter~6, pages 194--281. MIT Press,
  1986.

\bibitem[Srivastava et~al.(2014)Srivastava, Hinton, Krizhevsky, Sutskever, and
  Salakhutdinov]{Srivastava14}
N.~Srivastava, G.~Hinton, A.~Krizhevsky, I.~Sutskever, and R.~Salakhutdinov.
\newblock Dropout: A simple way to prevent neural networks from overfitting.
\newblock \emph{Journal of Machine Learning Research}, 15:\penalty0 1929--1958,
  2014.

\bibitem[Tsallis(2009)]{Tsallis09}
C.~Tsallis.
\newblock \emph{Introduction to Nonextensive Statistical Mechanics: Approaching
  a Complex World}.
\newblock Springer New York,, 2009.

\end{thebibliography}

\clearpage

\begin{appendices}
\section{Software and implementation.}
\label{Software and}
The optimizer used is Adam  \cite{Kingma14-2} stochastic gradient descent back-propagation. Specific hyper-parameters are in the text. We used only two standard sets of hyper-parameters, one for the classifier branch and one for the auto-encoder branch, no optimizations. 

For reconstruction error of a binarized $\mu$-th  row-observation $\mathbf{x}_{\mu}$ $=\{x_{\mu i}\}_{i=1}^N$ and its reconstruction $\mathbf{\hat{x}}_{\mu}$, we use the standard binary cross-entropy  \cite{Theano12}: $\mathbb{E}_{\mathbf{x}_{\mu}}[- \log \mathbf{\hat{x}}_{\mu}]$ $ =  \sum_i (- x_{\mu i}\log \hat{x}_{\mu i} $ $ -(1-x_{\mu i}) (1- \log \hat{x}_{\mu i}) )$, using a sigmoid last non-linearity $\varphi$. The batch cross-entropy is $-\log \mathcal{L}_{recon}$ $=\frac{1}{B}\sum_{\mu=1}^B$ $\mathbb{E}_{\mathbf{x}_{\mu}}[- \log \mathbf{\hat{x}}_{\mu}]$. In the space of observations, as on right plot of Figure \ref{Fig4.1}, the dual reconstruction error is the same binary cross-entropy $\mathbb{E}_{\mathbf{x}_{i}}[- \log \mathbf{\hat{x}}_{i}]$, but for the $i$-th observable   $\mathbf{x}_{i}$ $=\{x_{\mu i}\}_{\mu=1}^B$ and a sum over $\mu$ instead of $i$. The batch cross-entropy is $-\log \mathcal{L}'_{recon}$ $=\frac{1}{B}\sum_{i=1}^N$ $\mathbb{E}_{\mathbf{x}_{i}}[- \log \mathbf{\hat{x}}_{i}]$, with a normalization factor conforming to the space of observables.

The non-linearities are $\tanh()$ in the auto-encoder branch and two-unit \emph{maxout} \cite{Goodfellow13} in the classifier branch. Weight matrices of size $P \times N$ are initialized as random Gaussian matrices, normalized by the order of magnitude $\sqrt{P} + \sqrt{N}$ of their largest eigen-value. As discussed in sub-section \ref{Non-generative ACE}, hidden observables in the first and last hidden layer of  classifiers are batch-normalized i.e. de-meaned and divided by their second moment. Unlike  \cite{Ioffe15}, batch normalization is enforced identically in both the train and test set, hence  test results depend slightly on the test batch.

\section{Generative error formulas.}
\label{Laplacian and}
When sampling from a Laplacian, in order to have an unity variance in the prior, we choose for the independent one-dimensional latents a prior $p(z)$ $=p^{Lap}(z; 0,\sqrt{0.5})$, where $p^{Lap}(z; \mu,b)$ $=exp(-|z-\mu|/b)/(2b)$ is the standard Laplacian density with mean $\mu$ and scale $b$. In order to have zero generative error when $(\mu, \sigma)$ $\rightarrow (0,1)$, we parametrize the conditional posterior as $p(z|.)$ $=p^{Lap}(z; \mu,\sigma \sqrt{0.5})$. The generative error in (\ref{3.6}) equals: $- \log \sigma$ $ + |\mu|/\sqrt{0.5}$ $ + \sigma exp\left(- |\mu|/(\sigma\sqrt{0.5})\right ) - 1$,  see \cite{Gil13}, table 3.

For the divergence between a mixture prior $\sum_s\alpha_s p_s(z)$ and a mixture posterior $\sum_s\alpha_s p_s(z| .)$ with the same weights $\{\alpha_s \}_s$, we use the upper bound $\sum_s\alpha_s$ $ \mathcal{D}\left(p_s(z| .) || p_s(z) \right) $ implied by the log sum inequality \cite{Cover06}. For improvements, see open problem \ref{Use a} in section \ref{Open problems}.
\end{appendices}

\end{document}